\pdfoutput=1

\documentclass{article}
\usepackage{arxiv}
\usepackage{cite}
\usepackage{amsmath,amssymb,amsfonts}
\usepackage{algorithmic}
\usepackage{graphicx}
\usepackage{textcomp}
\usepackage{multirow,booktabs}
\usepackage{xspace}
\usepackage{graphicx} 
\usepackage{algorithmic}
\usepackage{siunitx}
\usepackage{gensymb}
\usepackage{CJKutf8}
\usepackage{times}    
\usepackage[labelfont=bf, textfont=bf]{caption}

\usepackage{bm}
\makeatletter
\AtBeginDocument{\DeclareMathVersion{bold}
\SetSymbolFont{operators}{bold}{T1}{times}{b}{n}
\SetMathAlphabet{\mathrm}{bold}{T1}{times}{b}{n}
\SetMathAlphabet{\mathit}{bold}{T1}{times}{b}{it}
\SetMathAlphabet{\mathbf}{bold}{T1}{times}{b}{n}
\SetMathAlphabet{\mathtt}{bold}{OT1}{pcr}{b}{n}
\SetSymbolFont{symbols}{bold}{OMS}{cmsy}{b}{n}
\renewcommand\boldmath{\@nomath\boldmath\mathversion{bold}}}
\makeatother

\def\BibTeX{{\rm B\kern-.05em{\sc i\kern-.025em b}\kern-.08em
    T\kern-.1667em\lower.7ex\hbox{E}\kern-.125emX}}

\begin{document}
\begin{CJK}{UTF8}{gbsn}

\title{Ubiquitous Field Transportation Robots with Robust Wheel-Leg Transformable Modules}

\newcommand{\orcid}[1]{\textsuperscript{#1}}

\author{
    \textbf{Haoran Wang}\orcid{1}, 
    \textbf{Cunxi Dai}\orcid{1,2}, 
    \textbf{Siyuan Wang}\orcid{1}, 
    \textbf{Ximan Zhang}\orcid{1}, 
    \textbf{Zheng Zhu}\orcid{1},\\
    \textbf{Xiaohan Liu}\orcid{1,2}, 
    \textbf{Jianxiang Zhou}\orcid{1,3}, 
    \textbf{Zhengtao Liu}\orcid{1,4}, 
    \textbf{Zhenzhong Jia}\orcid{1}
\thanks{Corresponding author: Zhenzhong Jia (e-mail: jiazz@sustech.edu.cn).}
}
\vspace{-10pt}
\maketitle

\noindent\orcid{1} Department of Mechanical and Energy Engineering, Southern University of Science and Technology, China. \\
\orcid{2} Now at Robotics Institute, Carnegie Mellon University, Pittsburgh, USA. \\
\orcid{3} Now at Hong Kong University of Science and Technology, China. \\
\orcid{4} Now at National University of Singapore, Singapore.

\footnotetext{This work was supported by the Science, Technology and Innovation Commission of Shenzhen Municipality (Grant No. KCXFZ20201221173006016, ZDSYS20220527171403009); Guangdong Province Science and Technology Program (2021QN02Z560); Guangdong Natural Science Fund-General Programme (2021A1515012384); and the National Science Foundation of China (NSFC) \#U1913603.}
\vspace{10pt}


\begin{abstract}
This paper introduces two field transportation robots. Both robots are equipped with transformable wheel-leg modules, which can smoothly switch between operation modes and can work in various challenging terrains. 
SWhegPro, with six S-shaped legs, enables transporting loads in challenging uneven outdoor terrains. SWhegPro3, featuring four three-impeller wheels, has surprising stair-climbing performance in indoor scenarios. 
Different from ordinary gear-driven transformable mechanisms, the modular wheels we designed driven by self-locking electric push rods can switch modes accurately and stably with high loads, significantly improving the load capacity of the robot in leg mode. 
This study analyzes the robot's wheel-leg module operation when the terrain parameters change. 
Through the derivation of mathematical models and calculations based on simplified kinematic models, a method for optimizing the robot parameters and wheel-leg structure parameters is finally proposed.
The design and control strategy are then verified through simulations and field experiments in various complex terrains, and the working performance of the two field transportation robots is calculated and analyzed by recording sensor data and proposing evaluation methods.

\end{abstract}


\section{Introduction}
Wheeled platforms excel in traversing flat surfaces with high speed and smooth motion with great energy efficiency; however, they often encounter challenges such as slippage and immobility over irregular and challenging terrains~\cite{kim2014wheel}. 
In contrast, legged robots demonstrate superior adaptability over uneven terrains; however, their cost is reduced speed and efficiency when compared to wheeled platforms on flat ground~\cite{cao2022omniwheg}. 
Addressing this trade-off is where the conceptualization of a hybrid wheel-leg transformable robot comes in, aiming to optimize performance by using wheeled mode for flat environments and legged mode for challenging terrains~\cite{chen2014Quattroped, sun2017transformable}. 
This strategic integration, which seeks a versatile and efficient locomotion system across diverse environmental conditions, has been adopted by many mobile robotic platforms.

\begin{figure*}[t!]
    \centering
    \includegraphics[width=0.95\linewidth, scale=0.8]{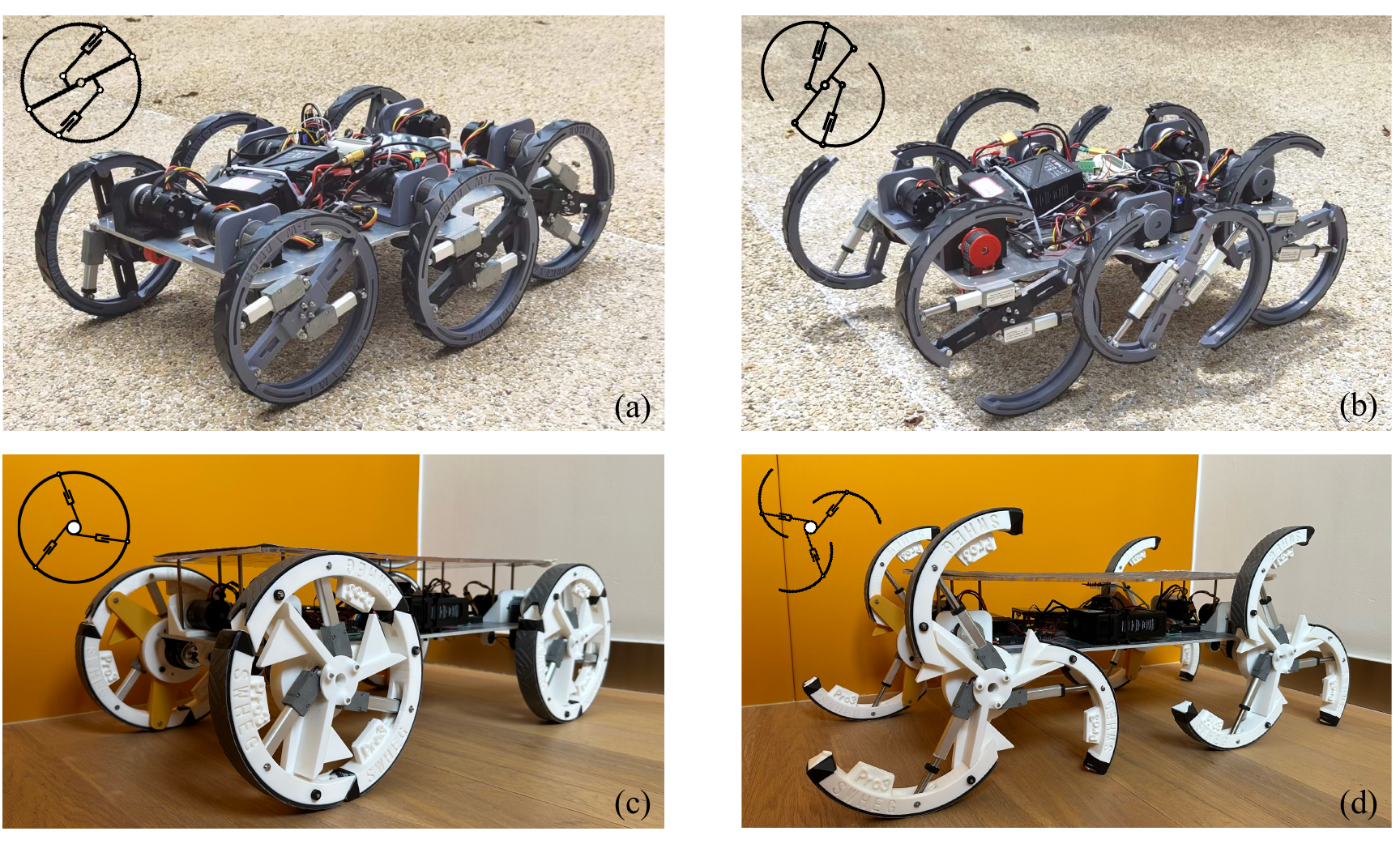}
    \vspace{-3mm}
    \caption{The two hybrid wheel-leg transformable robots developed in this paper, with schematic diagram of the wheel-leg module operational mode shown in the top left corner of each sub-figure.
    Top row: the SWhegPro robot has 6 wheel-leg transformable modules, which can be switched between wheel mode (a) and S-shaped leg mode (b). 
    Bottom row: the SWhegPro3 robot has 4 wheel-leg transformable modules, which can be switched between wheel mode (c) and three-impeller leg mode (d). 
    The mode switching of each wheel-leg transformable module is achieved by the self-locking electric push rods, which can significantly improve the robot's robustness and load capacity.
    }
    \vspace{-2mm}
    \label{fig:SWhegPro&Pro3}
    \centering
\end{figure*}

The concept of hybrid wheel-leg mobile robots is relatively broad; many robots with different hybrid wheel-leg realizations have shown improvements in mobility and energy efficiency. 
For instance, heteromorphic wheel mechanism usually features specialized fixed-shape wheels, such as C-shaped wheels for RHex robot \cite{Bai2018An,Marques2006RAPOSA,Saranli2001RHex,agrawal2016ions}. 
Another type of robot has separate wheel and leg mechanisms coexisting on the platform, and the collaboration of these two mechanisms generate locomotion. For instance, Chariot III has two big wheels and four $3$-DOF (degree-of-freedom) legs \cite{Nakajima2004Motion}. 
PEOPLER-II has two bars mounted on each of its four wheels, and its locomotion can be switched between leg mode and wheel mode\cite{Kosugi1984Motion}. Wheeleg has two pneumatically actuated $3$-DOF front legs and two independently driven rear wheels \cite{Lacagnina2003Kinematics}.

Wheel-leg transformable robots differ vastly from the transformable mechanism design to application scenario perspectives. 
First, from the morphology perspective,
wheel-leg transformable robots distinguish themselves through adaptive morphological alterations between wheel and leg configurations. 
Certain instances incorporate passive 1-DOF mechanisms per wheel-leg module, effecting transformation solely upon encountering a vertical obstacle plane. Notable exemplars encompass the wheel transformer, land devil ray, and shape-morphing wheel~\cite{WLinARM,Bai2018An,ryu2020shape}. 
Conversely, active 1-DOF mechanisms, typified by origami-based wheel transformation~\cite{Lee2017Origami}, enable nuanced adaptations, such as the robotic wheel overcoming obstacles via spoke manipulation~\cite{Moriya2020Robotic}. 
FUHAR, in a sophisticated maneuver, deploys a finger-like spoke for mode switching during stair negotiation~\cite{R2020FUHAR}. 
Recognizing limitations inherent to 1-DOF mechanisms in accommodating a diverse spectrum of obstacle sizes, advanced exploration has transpired toward 2-DOF mechanisms. 
Platforms such as Quattroped and STEP exemplify this paradigm, employing 2-DOF mechanisms for nuanced mode transformations. 
Specifically, Quattroped integrates a WHEG (WHeel-lEG) mechanism facilitating the transition between the wheel and S-shaped leg configurations, strategically optimizing obstacle traversals~\cite{chen2013quattroped,kim2020step}.

From the application scenario perspective,
building upon the general ability to overcome challenging terrains by changing into different shapes, previous studies have customized the wheel shapes to suit both outdoor exploration scenarios and indoor service scenarios. 

For instance, Won \textit{et al.} designed a 2-DoF wheel-leg transformable module that transforms into three-spoke wheels and excels in indoor stair climbing~\cite{won_design_2022}.
The spoke number is a noticeable difference between wheel-leg transformable robots designed to work outdoors and indoors. 
The outdoor transformable robots mostly have two spokes that bring better terrain clearance, while indoor transformable robots usually have three spokes to better suit the common stair-climbing tasks.

Previously, our group has developed a novel wheel-leg transformable robot, SWheg, with minimalist actuator realization~\cite{dai2022swheg}. 
The unique tendon-driven transformation mechanism allows a central servo to switch the operation modes of multiple wheel-leg transformation modules simultaneously. 
Hence, the same actuation system can be used in both wheeled and legged modes.
However, the tendon-drive system increases system complexity and requires delicate calibration. 
It is less robust, especially when dealing with relatively large payload, thereby might requiring frequent repairing.


To mitigate these problems, we introduce new wheel-leg transformation modules featuring self-locking push rods for operation mode switching. 
As shown in Fig.~\ref{fig:SWhegPro&Pro3}, 
this paper introduces a series of robust wheel-leg transformable robots to fulfill different transportation tasks: SWhegPro~\cite{dai2022swhegpro}, with six 2-spoke wheel-leg modules, is designed to tackle challenging outdoor terrains; SWhegPro3~\cite{wang2023swhegpro3}, with four three-impeller wheel-leg modules, specializes for indoor stair-climbing scenarios.
The new wheel-leg transformation mechanism design has following advantages: 1) the transformation mechanism forms a stable triangular structure to support the rim; 2) the on-wheel electric push rods allow the platform to operate with a much higher payload; and 3) allow different rims on the same module to expand separately, resulting in more possible configurations and gaits.


In the subsequent sections, we will elaborate on the design principles and explain the rationale behind the choice of wheel morphology and the incorporation of a self-locking electric push rod. 
We will also provide detailed analyses of complicated motion on grounds and stairs, including the derivation of models for optimizing wheel-leg module configurations and push rod displacements to enhance their performance. 
Our proposed design and control strategies will be validated through simulation and on-site experiments.


\section{Hybrid Wheel-Leg Module Design}

\subsection{Design Principle of the Transformable Wheels}
\begin{figure*}[t!]
    \centering
    \includegraphics[width=\linewidth, scale=0.8]{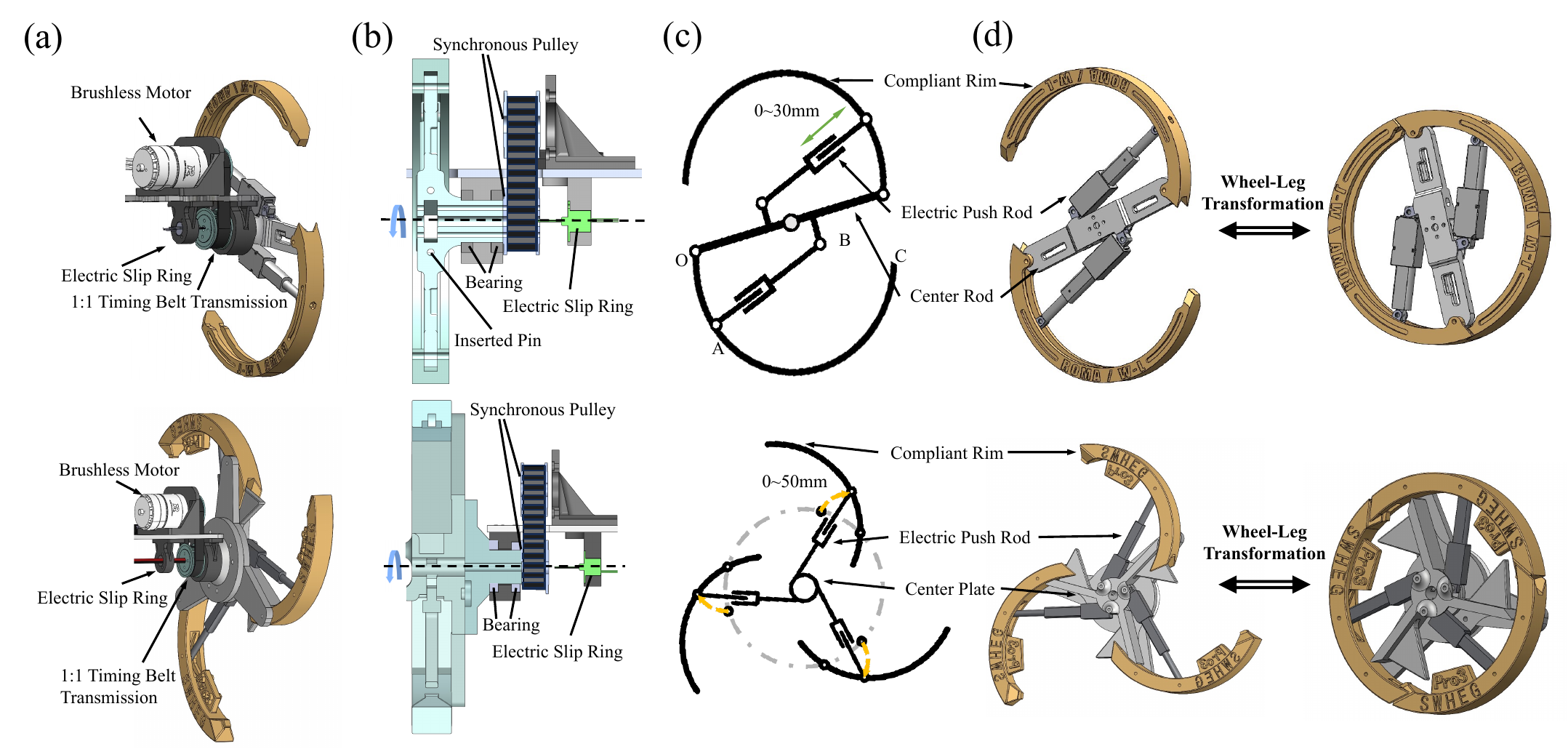}
    \vspace{-1mm}
  
    \caption{Detailed mechanism design of wheel-leg transformable modules, where the upper column is the module in SWhegPro and the lower column shows the module in SWhegPro3. The configuration of the module in legged modes (a). The section view of the transformation mechanism (b), showing the electric slip rings and the synchronous pulley. The wire-frame drawing (c) and the transformation between wheel mode and leg mode (d).}
    \vspace{-2mm}
    \label{fig:Detailed Mechanism Design}
    \centering
\end{figure*}

Based on the design of other wheel-leg robot, we think that reducing the complexity of transmission mechanism can make the wheel module self-sufficient and robust.
Therefore, we integrate the transformation actuators and the wheel together, directly switching the modes of wheels and legs.
In the wheel-leg module of SWheg robot\cite{dai2022swheg} we designed previously, the servos drive the ropes and switch the modes.
Its advantages are that the ropes occupy little space and the speed of mode switching is fast. However, the servos' torque is insufficient, which may leads the failure of transformation.
Thus, to improve the load capacity and stability of the robot, we decide to choose the electric push rod as the transformation actuator. 
Which is self-locked in any position within its operating range and cannot be driven in the opposite direction, significantly improving the stability of the wheel module.
According to experience in SWheg\cite{dai2022swheg}, independent wheel-leg modules reduce the operating cost and avoid disassembling the wheels for each maintenance.
Since each actuator requires an independent power line, entanglement may occur when the wheel rotates. Therefore, we use the electric slip ring to transmit electrical signals on the wheel and to ensure the power supplement is stable, which is shown in Fig.~\ref{fig:Detailed Mechanism Design}(a) (b).
Brushless motors with 39:1 planetary reducer are also a part of the module, and the transmission ratio of the synchronous belt is 1:1, which both are shown in Fig.~\ref{fig:Detailed Mechanism Design}(a).

\subsection{Parameters of the Wheels}
For SWhegPro and SWhegPro3, the geometric parameters of the modular wheel should be selected according to their target working scenarios. This part shows the parameters selection process of the robots.

\textbf{SWhegPro's parameters:} 
Aiming to enable SWhegPro to transport things more stable in various uneven outdoor scenes, we choose S-shaped wheels for the leg mode.
This is similar to its predecessors SWheg\cite{dai2022swheg} and Turboquad\cite{chen2017turboquad}, which show outstanding performance in balancing their ability of terrain adaptation and motion smoothness of the platform. 
To achieve the design goal of being robust, the transformation mechanism is designed to be a stable triangular structure with a center rod, electric push rod, and base.
The compliant areas in S-shaped legs improve the robot platform's energy efficiency and motion stability. 
Therefore, our goal is to maximize the compliant area of the rim in the leg, i.e., the rim AC labeled in Fig.~\ref{fig:Detailed Mechanism Design}(c).

We construct a geometric model of the wheel in MATLAB to select parameters for the transformation mechanism and to achieve two goals: maximize the compliant area in the rim and ensure the leg edges will not cause interference.

\textbf{SWhegPro3's parameters:} 
Our goal is to transport things more stable in indoor scenes with stairs of different sizes.
Inspired by curved-spoke tri-wheel robot\cite{kim2019curved} and WheelLeR\cite{8793686}, we believe the three-impeller structure has excellent potential for stair climbing. Therefore, we choose the curved-spoke wheel as the structure of the leg mode.

Based on the standards of civil stairs in China, we determine a collection of stairs within a specified size range, that is, length $240-300 mm $ and height $135-165 mm $.
Therefore, the geometric parameters of the curved-spoke structure are re-evaluated based on the original dimensions of the three-impeller.

\begin{figure}[t!]
    \centering
    \includegraphics[width=0.65\linewidth, scale=0.7]{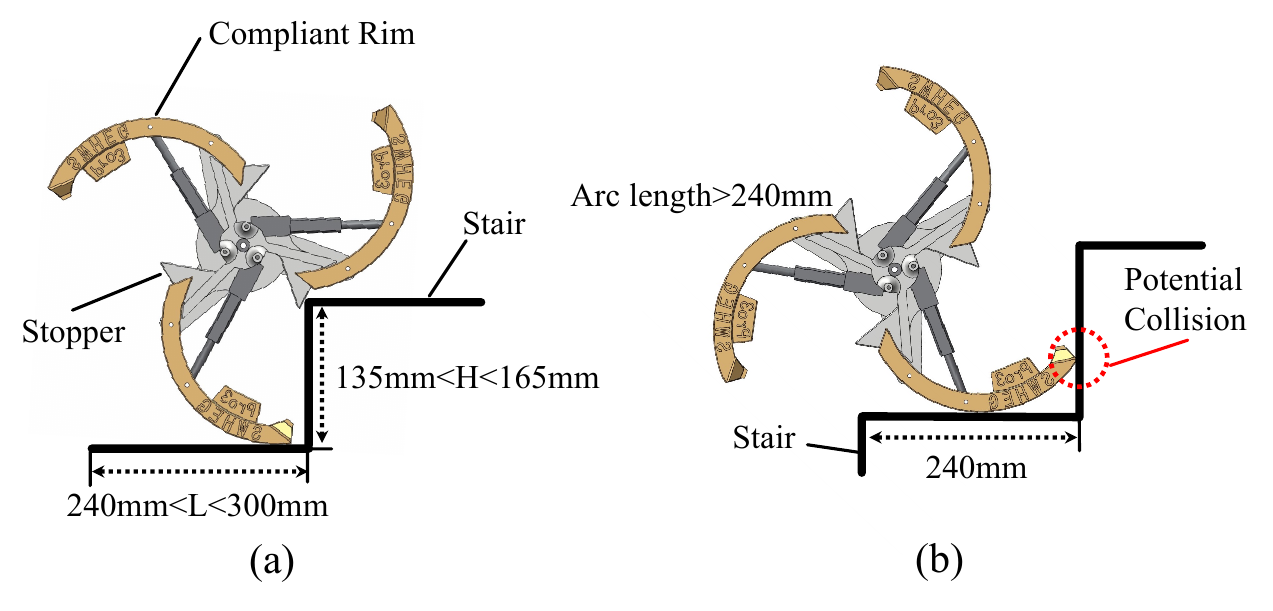}
    \vspace{-1mm}
    \caption{Situations should be considered to determine the wheel parameters. The contact position between the wheel and the standard stair is shown in (a). The limitation while meeting inappropriate stairs, e.g., the length of the rim is too long as (b) shown.}
    \vspace{-3mm}
    \label{Explain the Parameters of Wheel}
    \centering
\end{figure}

Through the kinematic analysis of a typical stair size $300mm*160mm$, we determine the minimum expand magnitude of the spoke in leg mode and the minimum length of an individual spoke. 
Through these demands, we reversely deduce the parameters of the wheel module in leg mode. The geometric demands are shown in Fig.~\ref{Explain the Parameters of Wheel}, which ensures that the spokes will not collide with the vertical surface of the stairs.
The stopper structure in the curved-spoke mechanism \cite{kim2019curved} is retained in SWhegPro3, whose reset function enhances the error tolerance during the climbing process.

Referring to the design parameters explained above, we have designed two transformation wheel modules, both of which are equipped with electric push rods to achieve smooth switching between wheel mode and leg mode. The geometric shapes of SWhegPro and SWhegPro3 are shown in Fig.~\ref{fig:Detailed Mechanism Design} (c).

\subsection{Wheels Number Selection}

Theoretically, the more wheels a robot has, the smoother its movement will be, but the number of wheels will affect the gait complexity and motion stability in leg mode.
Thus, we should compare the motion performance of the robots with different wheel numbers.
Based on experience gained from SWheg robot\cite{dai2022swheg} and curved-spoke robot\cite{kim2019curved}, we believe that a robot with four or six wheels is sufficient to meet the requirements. 
Then, we carried out a comparative examination of the robots that have four wheels or six wheels in the simulation.
An IMU module is placed at the center of the robot to collect the motion data, where a set of ZYX Euler represents the direction of the robot movement angles.
Therefore, the stability of the movement can be evaluated by the cost function $J(\phi, \theta, \psi)$, where $\psi$ is the yaw angle, $\theta$ is the pitch angle, and $\phi$ is the roll angle.

\begin{equation}
    J(\phi, \theta, \psi) = \sqrt{\phi^{2}+\theta^{2}+\psi^{2}}
\end{equation}

But for SWhegPro3, climbing parallel stairs makes the left and right sides the same, so the yaw angle($\psi$) and the roll angle($\phi$) equal 0.

\begin{figure}[t!]
    \centering
    \includegraphics[width=0.8\linewidth, scale=0.7]{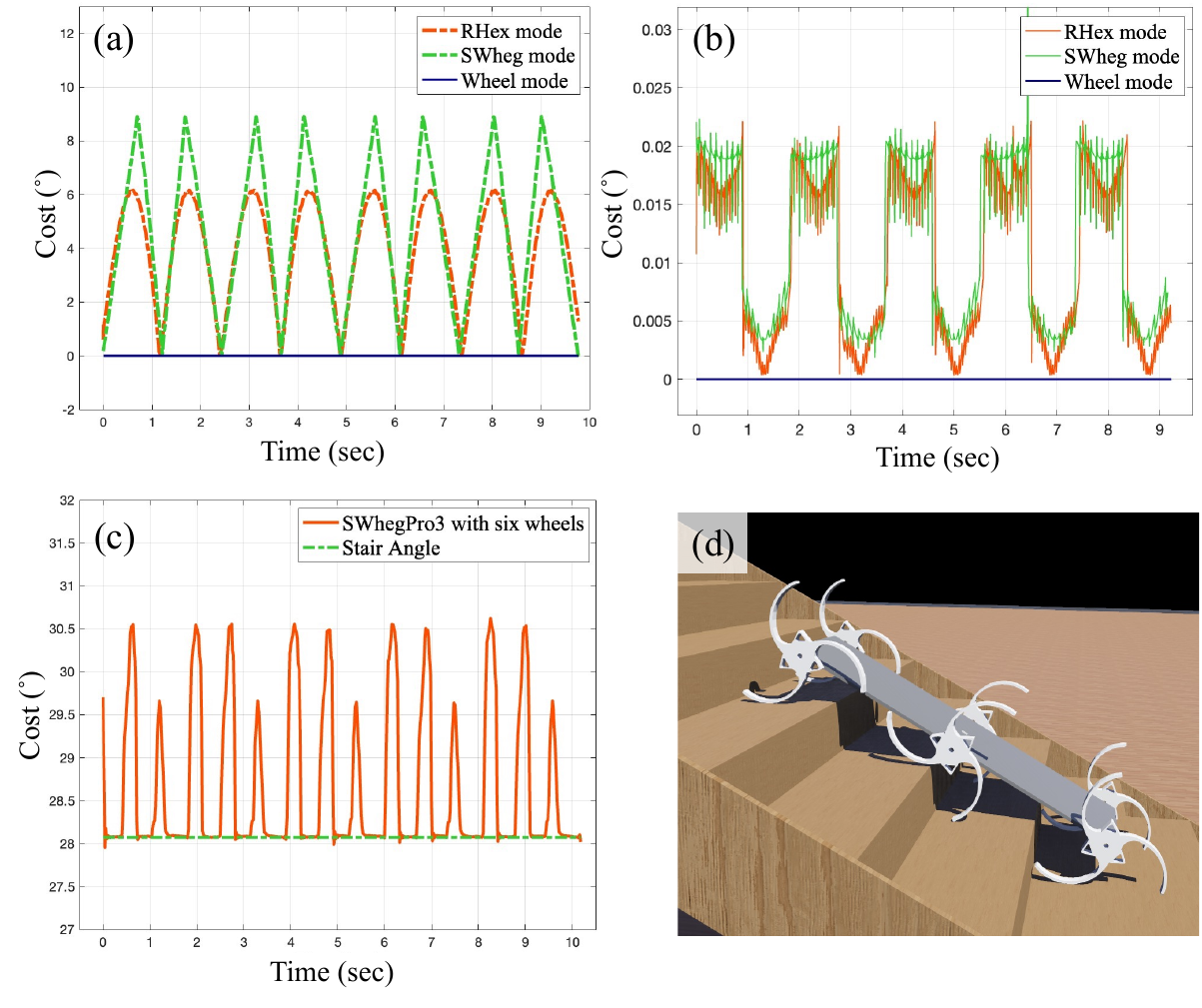}
    \vspace{-2mm}
    \caption{Comparison of cost when robots equipped with different number of wheels. The cost curves of SWhegPro runs on flat ground which equipped with 4 wheels (a) and 6 wheels (b). The cost curve of SWhegPro3 equipped with six wheels while climbing the stairs (c) and its schematic diagram in Webots simulation environment (d).}
    \vspace{-3mm}
    \label{Cost of Different Wheel Number}
    \centering
\end{figure}

It can be seen from Fig.~\ref{Cost of Different Wheel Number} that in leg mode, the gait of SWhegPro with only four wheels is volatile, the two-point contact structure lacks support; SWhegPro with six wheels supported by triangular structures, this gait can achieve a better stability.

Fig.~\ref{Cost of Different Wheel Number} shows that for SWhegPro3 with six wheels, the extra two legs will shake the robot at a high frequency, leading to unstable performance, and the robot could be too long than we expect to avoid wheels from interfering with each other.
The four-wheeled SWhegPro3 has no redundant structure, and its gait control strategy is more straightforward than that of the six-wheeler.

Thus, we believe that SWhegPro which mainly moves on the ground, six modular wheels can improve gait stability; 
Meanwhile, the six-wheeled design seems redundant for SWhegPro3, especially when climbing continuous stairs. The 4-wheel structure can climb stairs stably and efficiently.

\subsection{Wheelbase Identification}
\begin{figure}[h]
    \centering
    \includegraphics[width=0.55\linewidth, scale=0.8]{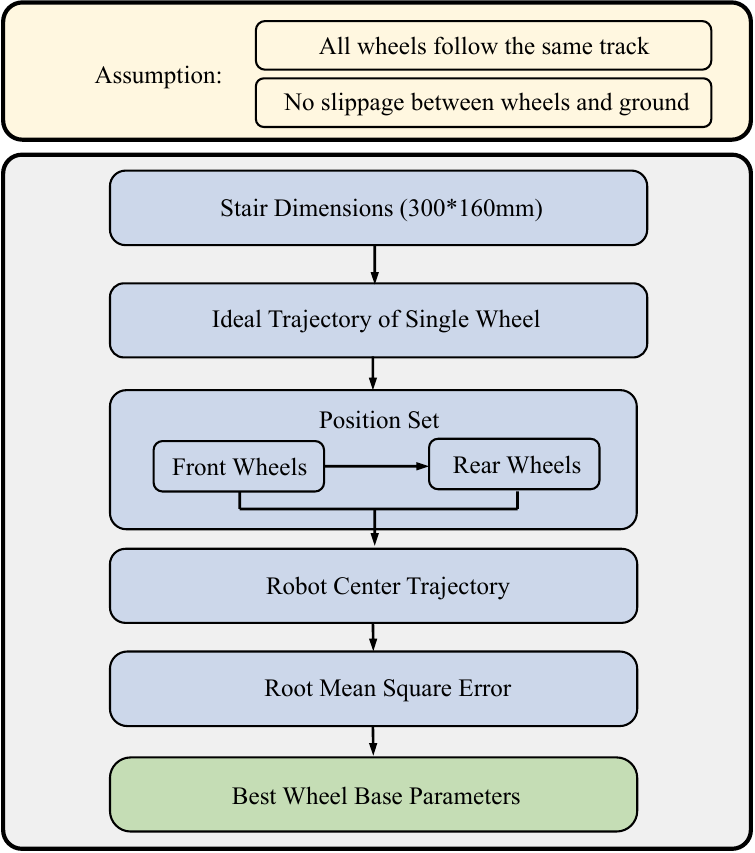}
    \vspace{-1mm}
    \caption{Assumptions and the flow diagram of optimal wheelbase calculation while designing SWhegPro3.}
    \vspace{-3mm}
    \label{Flow Diagram of Kinematics Analysis}
    \centering
\end{figure}

For SWhegPro, the distribution of the wheels only has to ensure no interference between the wheels in the leg mode. Therefore, the wheelbase is determined in proportion to the radius of the rim.

\begin{figure}[h]
 \centering
    \includegraphics[width=0.8\linewidth, scale=0.8]{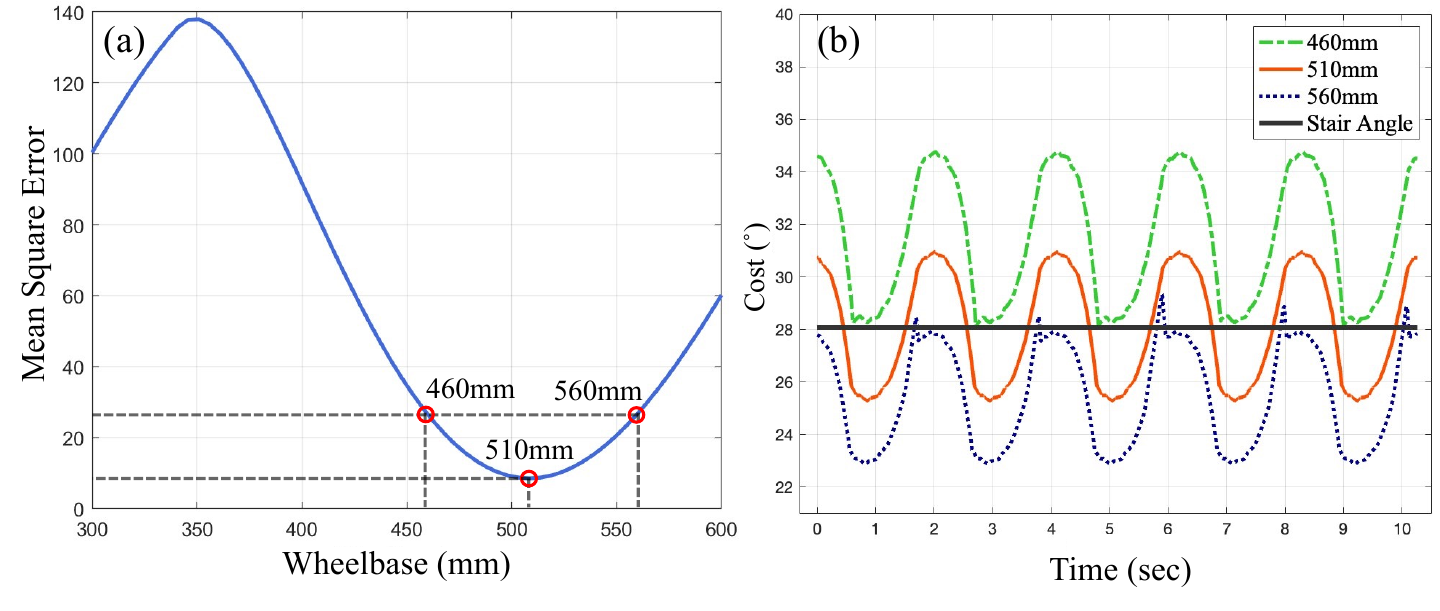}
    \vspace{-1mm}
    \caption{(a) shows the RMSE(Root Mean Square Error) value of SWhegPro3 while climbing at different wheelbases and (b) is the comparison of its cost at different wheelbases according to simulation.}
    \vspace{-1mm}
    \label{Center Error}
    \centering
\end{figure}

For SWhegPro3, what can be predicted is that there must be a phase difference sequence between the front and rear wheels while climbing the stairs.
Therefore, an appropriate wheelbase is key to enhancing the stability of SWhegPro3, so we use MATLAB to perform kinematic analysis to find the optimal solution.
We assume there is no slippage between the spokes and the stairs, and all wheels will follow the same ideal trajectory. The optimal wheelbase can be determined through some kinematic analysis and parameter optimization methods shown in Fig.~\ref{Flow Diagram of Kinematics Analysis}.
The optimization result is shown in Fig.~\ref{Center Error}(a). When the RMSE while moving is minimized, the corresponding wheelbase(510mm) is the optimal solution.

To verify the superiority of the wheelbase selection, we examine SWhegPro3 with two different wheelbases(460mm \& 560mm) in the simulation environment and measure the performance by cost function $J(\phi, \theta, \psi)$.
As shown in Fig.~\ref{Center Error}(b), the robot with optimal wheelbase has higher climbing stability, when the shorter wheelbase might lead the robot to fall backward, and the longer wheelbase will lead to a sudden change of pitch angle, which means the climbing process is not stable enough.

\section{Mechatronics System Design}
The main computing power of both robot models is the Raspberry Pi 4B embedded computer, which uses the Robot Operating System (ROS) for inter-module communication.
DJI M3508 brushless reduction motors are used as the drive unit, and each motor is connected to a DJI C620 FOC controller.
The socket converts USB signals to CAN signals, and the CAN board connects the FOC to the embedded computer.
We use 12 electric push rods as wheel leg change actuators in two robots, and they are controlled by PWM signals to extend the wheel spokes.
The wheelset itself is 3D printed from PLA material. Shock-absorbing cotton and PVC pads are added to the outer edges of the wheels to cushion impact and prevent slipping.

The specifications of the robots are summarized in Table~\ref{tab:ROBOT SPECIFICATIONS}.

\begin{table*}[]
\renewcommand\arraystretch{1.5}
\centering
\caption{ROBOT SPECIFICATIONS}
\label{tab:ROBOT SPECIFICATIONS}
\begin{tabular}{cccc}

\multicolumn{1}{c}{\textbf{Parameters}}& \textbf{Details}  & \textbf{SWhegPro}            & \textbf{SWhegPro3}           \\ \hline
\multirow{2}{*}{Length}   & Body                           & 0.50m                        & 0.59m                        \\
                          & Hip-to-hip                     & 0.21m                        & 0.51m                        \\ \hline
\multirow{2}{*}{Width}    & Body                           & 0.30m                        & 0.30m                        \\
                          & Leg-to-leg                     & 0.33m                        & 0.43m                        \\ \hline
\multirow{3}{*}{Height}   & Body                           & 0.15m                        & 0.25m                        \\
                          & Ground to hip (legged mode)    & 0.10m                        & 0.15m                        \\
                          & Ground clearance (legged mode) & 0.19m                        & 0.36m                        \\ \hline
\multirow{2}{*}{Radius}   & Wheel-leg (i.e., rim) radius   & 0.1m                         & 0.12m                        \\
                          & Maximum radius of wheel-leg    & 0.13m                        & 0.22m                        \\ \hline
\multirow{5}{*}{Weight}   & Total                          & 10.08kg                      & 17.65kg                      \\
                          & Body                           & 7.39kg                       & 12.38kg                      \\
                          & Wheel-leg module(each)         & 0.34kg                       & 1.15kg                       \\
                          & Battery                        & 0.67kg                       & 0.67kg                       \\ \hline
\multirow{2}{*}{Actuator} & Driving                        & DJI M3508 motor(×6)          & DJI M3508 motor(×4)          \\
                          & Transformation mechanism       & Electric Push Rod(x12)       & Electric Push Rod(x12)       \\ \hline
\multirow{2}{*}{Sensors}  & IMU                            & (×1)                         & (×1)                         \\
                          & Power measurement              & (×1)                         & (×0)                         \\ \hline
\multirow{2}{*}{Battery}  & DJI TB48                       & (×1)                         & (×1)                         \\
                          & Power bank                     & (×1)                         & (×1)                         \\ \hline
\end{tabular}
\end{table*}

\section{Control}
\subsection{Wheeled Mode Motion}

\begin{figure}[h]
    \centering
    \includegraphics[width=0.7\linewidth, scale=0.8]{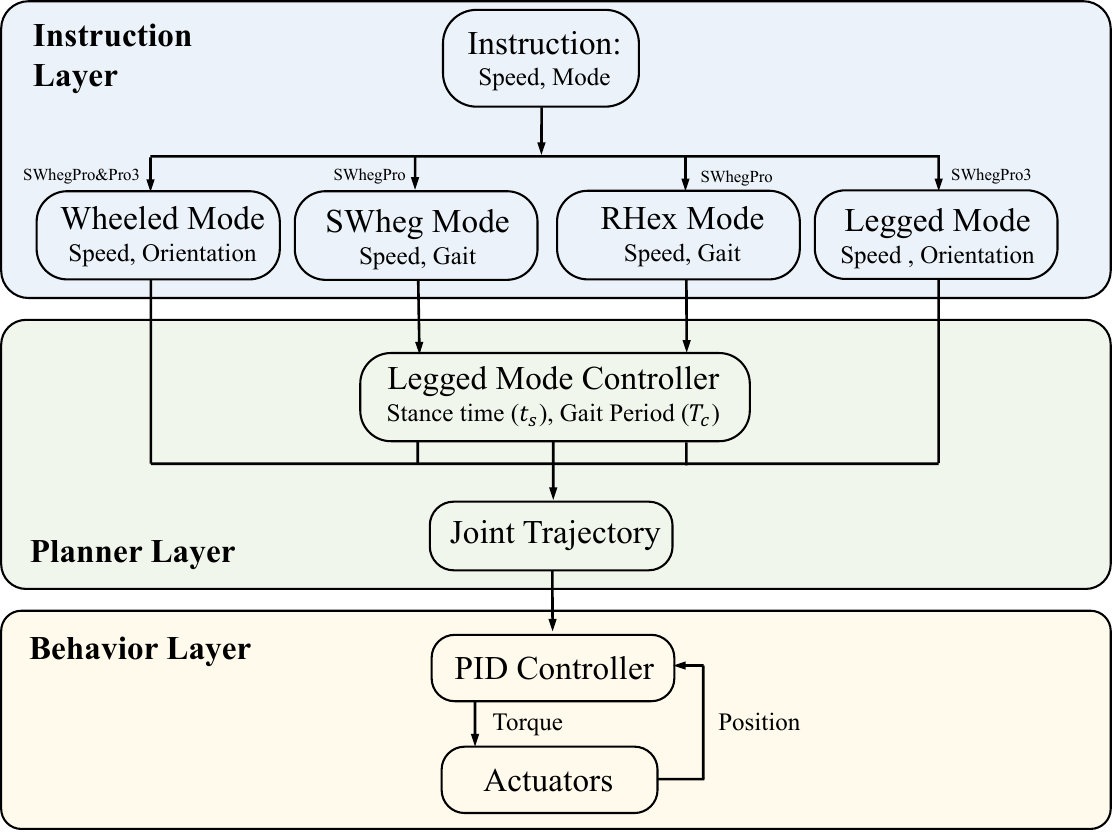}
    \vspace{-1mm}
    \caption{Overall control structure of the two robots.}
    \vspace{-2mm}
    \label{Control Structure}
    \centering
\end{figure}

The control system consists of three layers: the instruction layer, the planner layer, and the behavior layer. The instruction layer processes high-level instructions, including control mode, walking gait, and moving speed. Control modes are managed by a finite state machine. During mode transitions, the position and control mode of the joints will be adjusted accordingly. For now, the instructions are provided by a human operator. The desired speed, gait, and velocity are then sent to the controller of the corresponding mode. In the planner layer, the controller generates velocity/position trajectories for each joint according to the command. The behavior layer controls the actuators. The trajectory provided will be tracked by the actuators with PD controllers.Fig.~\ref{Control Structure}

When the robot operates in wheeled mode, the control of the joints is straightforward. Each joint's rotation velocity ($\omega_{i}$) is assigned based on the differential-wheel steering model. Joints on the same side of the robot will be given the same angular velocity, $\omega_{Left}$ or $\omega_{Right}$.

\begin{figure}[t!]
    \centering
    \includegraphics[width=0.45\linewidth, scale=0.7]{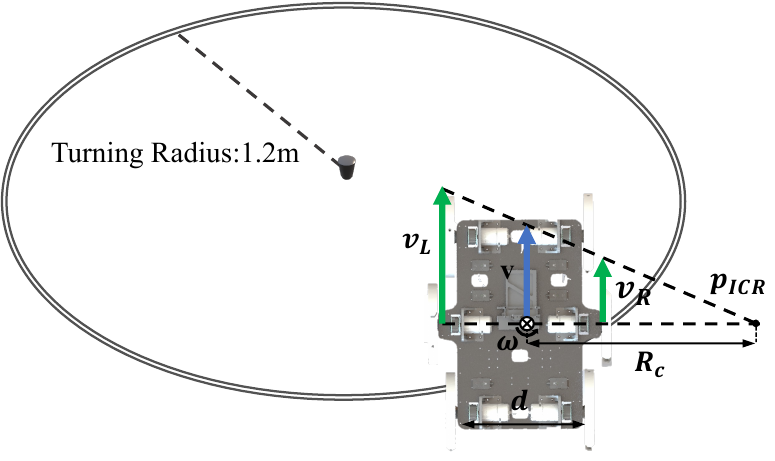}
    \vspace{-1mm}
    \caption{Turning radius of SWhegPro in wheeled mode on flat ground.}
    \vspace{-2mm}
    \label{Turning Radius}
    \centering
\end{figure}

\subsubsection{Legged Mode Motion}
\begin{figure}[t!]
    \centering
    \includegraphics[width=0.8\linewidth, scale=0.8]{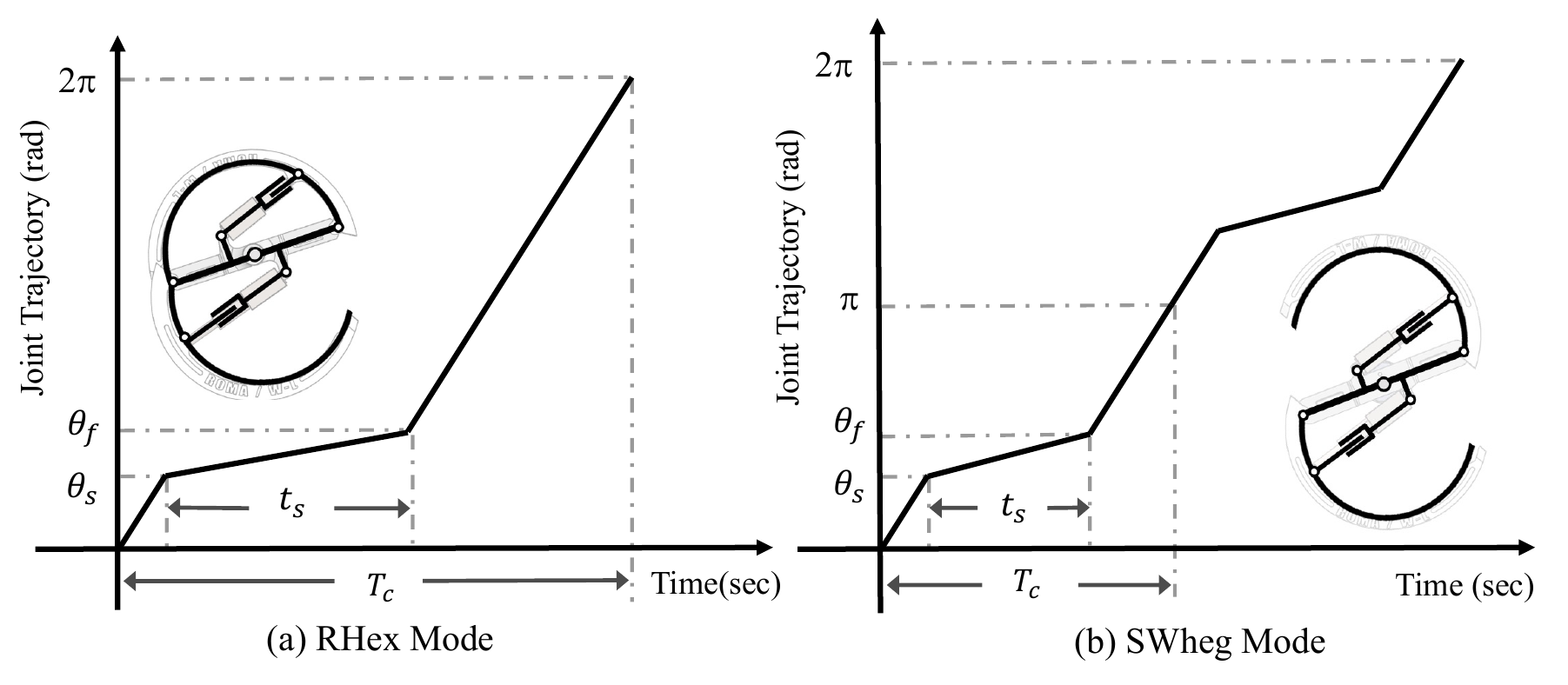}
    \vspace{-1mm}
    \caption{Joint trajectory of SWhegPro while operating on even terrains where (a) shows RHex mode and (b) shows SWheg mode.}
    \vspace{-2mm}
    \label{Joint Trajectory of Legged Modes}
    \centering
\end{figure}

As shown in Fig.~\ref{Turning Radius}, when the robot has a forward velocity $v_{desired}$ and a rotational velocity $\omega_{desired}$, its instant center of rotation will be located at point $P_{ICR}$. The Turning Radius can be expressed as:

\begin{equation}
R_{c} = \frac {v_{desired}} {\omega_{desired}}
\end{equation}
Then, the angular velocity of the joint on each side can be computed from
\begin{equation}
\omega_{Right} = \frac{ v_{desired} + \frac{\omega_{desired} \cdot d}{2}} {R_{wheel} N_{t}}\\
\end{equation}
\begin{equation}
\omega_{Left} = \frac{ v_{desired} - \frac{\omega_{desired} \cdot d}{2}} {R_{wheel} N_{t}} \\
\end{equation}

Where $d$ is the width of the chassis, and $N_t$ is the speed reduction ratio. The wheeled mode control strategy can also be used in robot platforms with different leg configurations.

\subsection{SWhegPro: Wheel-leg Transformation and Gait Strategies}

\subsubsection{Transformation Control}

\begin{figure*}[t!]
    \centering
    \includegraphics[width=0.95\linewidth, scale=0.8]{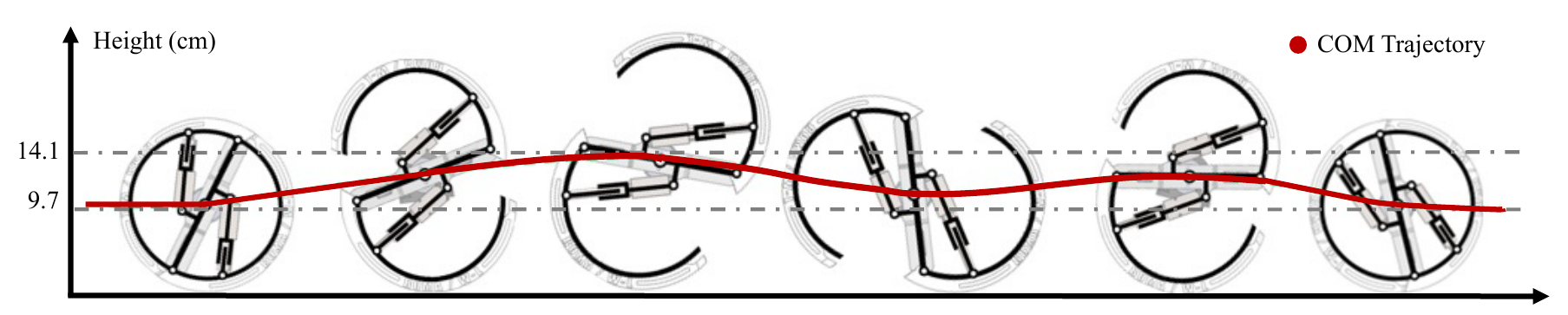}
    \vspace{-1mm}
    \caption{The center joint trajectory of SWhegPro's transformable module while operating on flat ground.}
    \vspace{-2mm}
    \label{The Module Center Joint Trajectory}
    \centering
\end{figure*}

\begin{figure*}[t!]
    \centering
    \includegraphics[width=0.95\linewidth, scale=0.8]{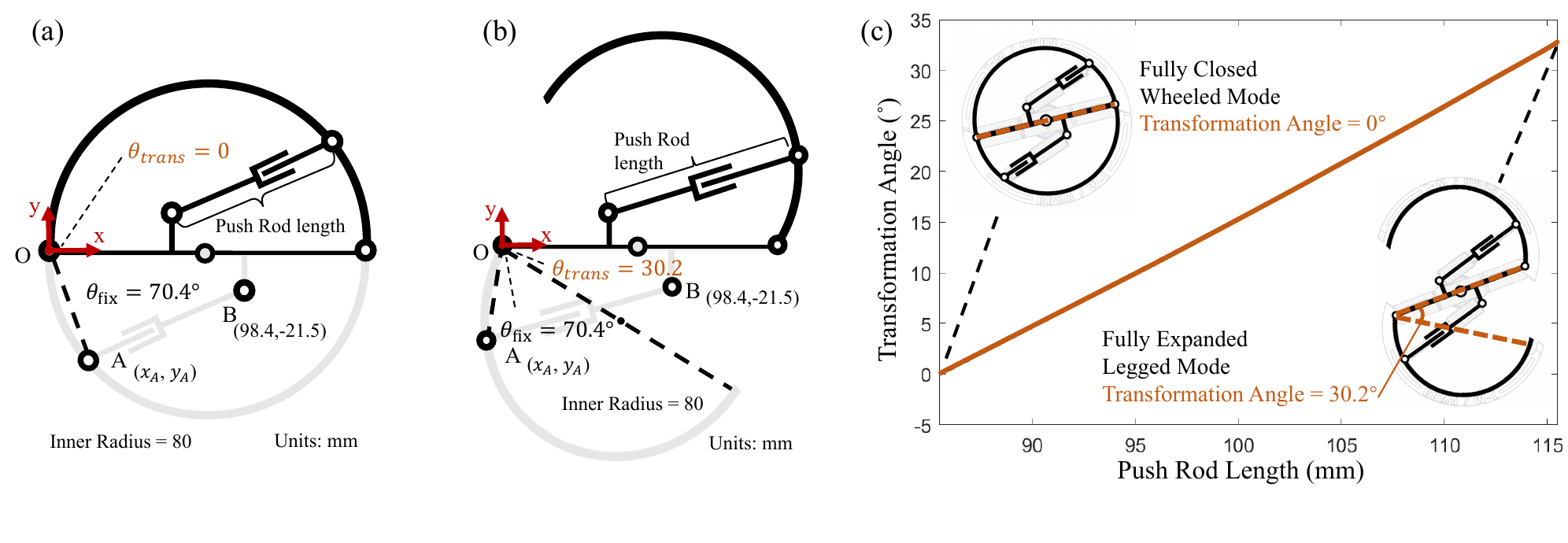}
    \vspace{-1mm}
    \caption{Linear mapping diagram of SWhegPro's modular wheel; the position of the rims when the module is fully closed (a) and fully opened (b); (c) is the linear mapping diagram between the length of the electric push rod and the angle of the rim.}
    \vspace{-2mm}
    \label{Kinematic Mapping of Pro}
    \centering
\end{figure*}

The change in the geometric center of the modular wheel during the transformation process is shown in Fig.~\ref{The Module Center Joint Trajectory}, which fluctuates between 10cm and 14.1cm.
To control the transformation of the wheel-leg module accurately, we simulate the kinematics of the transformation mechanism in MATLAB. A mapping between the length of the electric push rod and the transformation angle is used for control. 
The opening angle of the rim $\theta_{trans}$ and coordinates are shown in Fig.~\ref{Kinematic Mapping of Pro}(a)(b), and the linear kinematic mapping relationship is shown in Fig.~\ref{Kinematic Mapping of Pro}(b).

Referred to the calculation in SWheg robot\cite{dai2022swheg} we designed before, we find that the robot under SWheg mode has excellent stability and efficiency while walking on various complex, uneven ground, so it is not essential to enhance the motion performance by changing the opening angle of rim under leg mode.

\begin{equation}
   \left\{\begin{matrix} 
  \sqrt{{x_{A}}^2+{y_{A}}^2} =55 \\  
  \sqrt{({x_{A}-98.4)}^2+{(y_{A}+21.5)}^2} =l_{pushrod}\\
  y_{A}<0\\
\end{matrix}\right. 
\end{equation}

\begin{equation}
\theta _{trans} =
\left\{\begin{matrix} 
  \tan^{-1} \left | \frac{y_{A}}{x_{A}}  \right | - \theta _{fix} , &x_{A}\ge  0 \\  
  90 + \tan^{-1} \left | \frac{y_{A}}{x_{A}}  \right |  - \theta _{fix} &x_{A}< 0
\end{matrix}\right. 
\end{equation}

In legged modes, the planner-layer controller will generate a group of clock-driven periodic trajectories for joints. The trajectories generated are regular time functions, with outputs ranging from 0 to $2\pi$.

For RHex mode, the joint trajectory of walking mode has four key parameters: $\theta_{s}$, $\theta_{f}$, $t_s$, and $T_c$.(See Fig.~\ref{Joint Trajectory of Legged Modes} (a).) $T_c$ and $t_s$ determine the ground speed a Leg, where $T_c$ is the period of the function and $t_s$ is the approximate stance time. $\theta_{s}$ and $\theta_{f}$ are geometric parameters of the leg. We suppose that the SWhegs approximately hit the ground when $\theta = \theta_{s}$ and leave the ground when $\theta = \theta_{f}$. 

The joint trajectory of the SWheg mode is similar to that of the RHex mode. Due to the symmetrical design of the SWheg module, the joint trajectory of $ \pi < \theta < 2\pi$ repeats the trajectory of $ 0 < \theta < \pi$. (See Fig.~\ref{Joint Trajectory of Legged Modes} (b).)

By applying phase difference $t_{k}$ to the legs, different gaits can be generated. This naive gait-generation method can be applied to RHex and SWheg modes. The following table shows the phase difference of some simple gaits and the corresponding gait diagram. The definition of Wheel-Leg modules can be found in Fig.~\ref{Turning Radius} (b).

\begin{table}[h!]
\renewcommand\arraystretch{1.8}
  \begin{center}
    \caption{Phase Diagram}
    \begin{tabular}{ccccccc} 
      \textbf{Gait Type} & \textbf{RF} & \textbf{RM}  & \textbf{RR}  & \textbf{LF} & \textbf{LM}  & \textbf{LR}\\
      \hline
      tripod & 0 & $\frac{1}{2} T_{c}$ & 0 & $\frac{1}{2} T_{c}$ & 0 & $\frac{1}{2} T_{c}$ \\
      \hline
      ripple & 0 & $\frac{1}{3} T_{c}$ & $\frac{2}{3} T_{c}$ & $\frac{1}{6} T_{c}$ & $\frac{1}{2} T_{c}$ & $\frac{5}{6} T_{c}$ \\
    \end{tabular}
  \end{center}
  \vspace*{-5mm}
\end{table}

\subsection{SWhegPro3: Nonlinear gait and wheel module control}
\subsubsection{Motor Control System}

Unlike curved-spoke tri-wheel robot\cite{kim2019curved}, SWhegPro3 introduces a more complex control mode to drive the front and rear wheels together.
The driving wheels of SWhegPro3 can be divided into two groups: the front wheels are the main driving wheels, and the rear wheels are the auxiliary driving wheels.
This avoids the lack of kinetic energy output, but the cost is that the control strategy between the front and the rear wheels becomes more complicated.

\begin{figure}[t!]
    \centering
    \includegraphics[width=0.4\linewidth, scale=0.7]{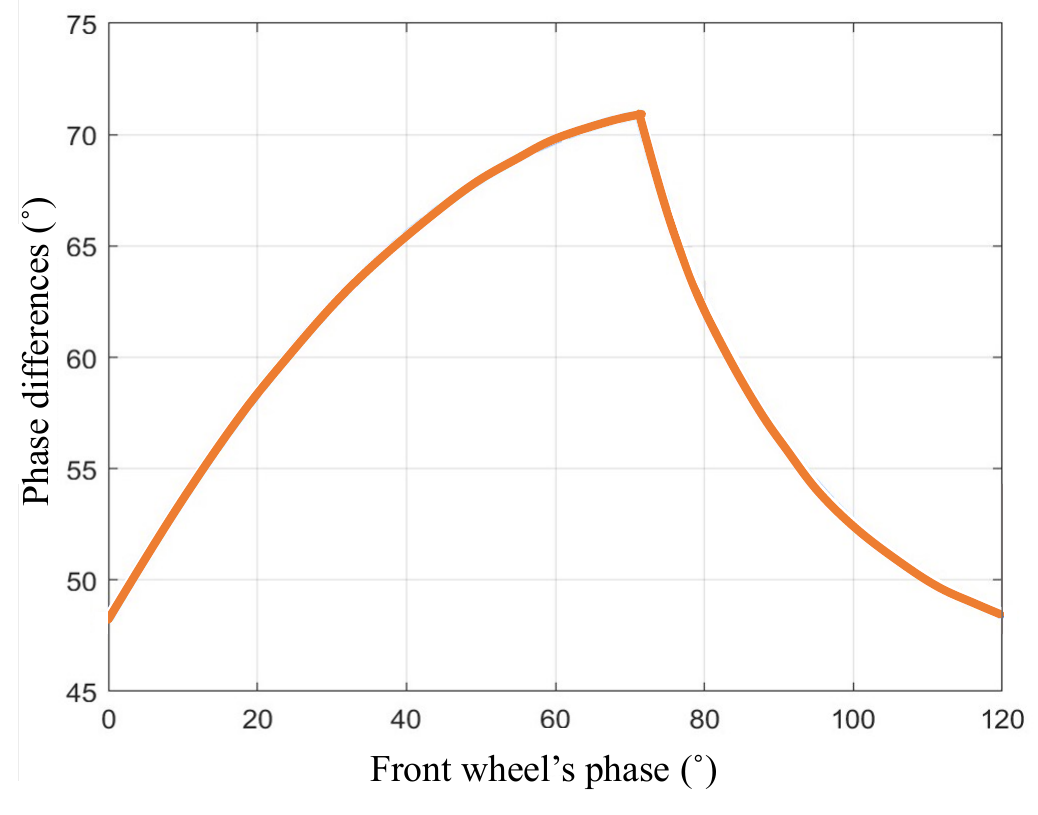}
    \vspace{-1mm}
    \caption{Dynamic phase difference between left and right wheels while SWhegPro3 climbing a stair dimensions of $L = 300mm$ and $H = 160mm$.}
    \vspace{-2mm}
    \label{Dynamic Phase Differences}
    \centering
\end{figure}

It can be expected that the phase between the front and rear wheels must not be a fixed value.
Therefore, we should obtain it in advance to control the climbing process by following the phase difference.
So, we used MATLAB to perform numerical simulations to capture the dynamic phase difference between the wheels, which is shown in Fig.~\ref{Dynamic Phase Differences}

Then, we fail to control the robot by following the dynamic phase difference.
The main reason is that dynamic phase difference is strictly related to the robot's position; we cannot locate the robot precisely during the actual situation, so the accumulative error will lead to unstable movement. 
Therefore, we excluded this overly idealized driving mode during the pre-experiment.

Then, we try to use the speed loop control, that is, making some compromises on control accuracy with the goal of stability.
We set torque output limits for the front and rear wheels to avoid excessive output when the accumulative error occurs.
Climbing performance testing and comparison are displayed in the Simulation and Experiment sections.

\subsubsection{Wheel Module Control}

\begin{figure*}[t!]
    \centering
    \includegraphics[width=0.95\linewidth, scale=0.8]{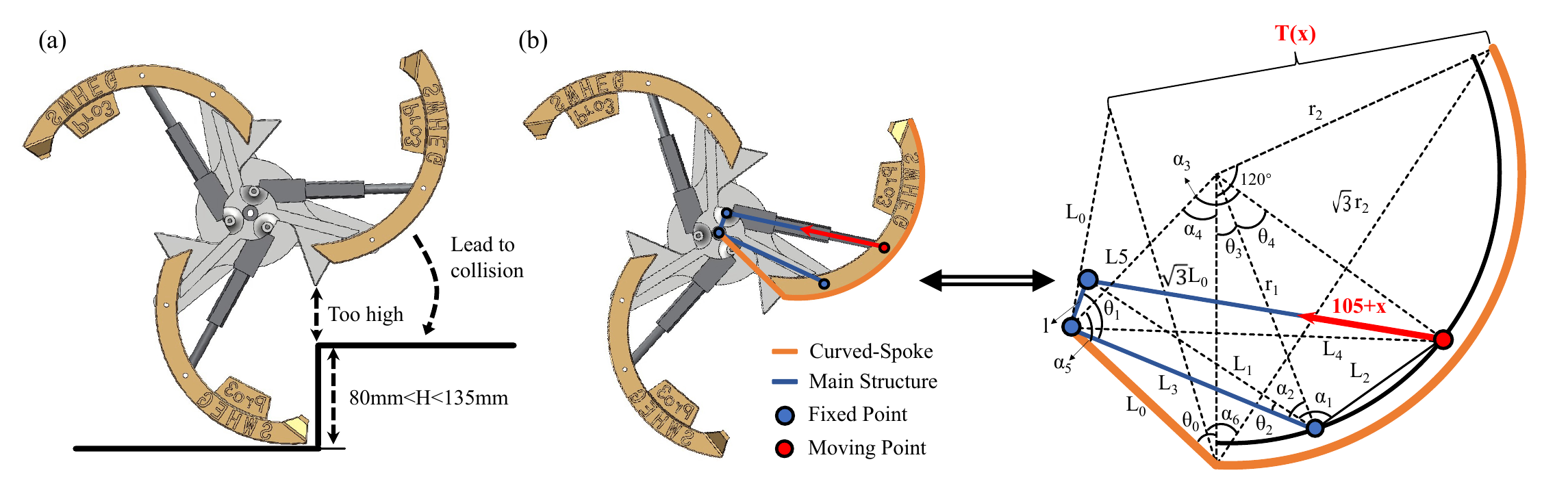}
    \vspace{-1mm}
    \caption{Kinematic mapping of SWhegPro3's modular wheel. (a) shows the condition when a fully opened wheel climbs a low stair. (b) indicates the relationship between the modular wheel and the mathematical model.}
    \vspace{-2mm}
    \label{Kinematic Mapping of Pro3}
    \centering
\end{figure*}

SWhegPro3 is designed to adapt to all dimensions of parallel stairs, whereas Fig~\ref{Kinematic Mapping of Pro3}(a) shows an unstable condition. 
So we explored a method: Dynamically adjusting the wheel shape by changing the angle of rims while climbing different sizes of stairs.
It uses intuitive geometric analysis to find the mapping between the length of the electric push rods and the parameters of the stairs.
In Fig.~\ref{Kinematic Mapping of Pro3}(b), the green lines represent the redesigned spoke curves, while the blue lines represent the main structural elements of the physical object. The red arrow represents the length of the electric push rod $x$, and $T(x)$ is the dependent variable to be solved.
The additional black dotted lines in Fig.~\ref{Kinematic Mapping of Pro3}(c) are auxiliary parameters.

Although the length of the electric push rods is constantly changing, some quantities are constant, among which:
$\theta_{0}=\ang{50}$, $\theta_{1}=\ang{92.72}$, $\theta_{2}=\ang{9.86}$, 
$\theta_{3}=\ang{21.53}$,
$\theta_{4}=\ang{34.05}$, 
$r_{1}=114mm$,
$r_{2}=124mm$, 
$l=20mm$,
$L_{1}=116.67mm$, 
$L_{2}=66.76mm$, 
$L_{3}=114mm$,
$L_{Arc}=240mm$.

Following are the analytic geometry derivation formulas and all the unknowns are shown in Fig.~\ref{Kinematic Mapping of Pro3}(c).

\begin{equation}\label{eq2}
    \alpha_{1}=\arccos(\frac{L_{1}^2+L_{2}^2-(105+x)^2}{2L_{1}L_{2}})
\end{equation}

\begin{equation}
    L_{4}=\sqrt{L_{2}^2+L_{3}^2-\cos{(\alpha_{1}+\theta_{2})*2L_{2}L_{3}}}
\end{equation}

\begin{equation}
    \alpha_{2}=\alpha_{1}+\theta_{2}-\frac{(\ang{180}-\theta_{4})}{2}
\end{equation}

\begin{equation}
    L_{5}=\sqrt{r_{1}^2+L_{3}^2-\cos{\alpha_{2}}*2r_{1}L_{3}}
\end{equation}

\begin{equation}
    \alpha_{3}=\arccos{(\frac{L_{5}^2+r_{1}^2-L_{4}^2}{2r_{1}L_{5}})}
\end{equation}

\begin{equation}
    \alpha_{4}=\alpha_{3}-\theta_{3}-\theta_{4} ; \alpha_{6}=\ang{180}-\alpha_{5}-\alpha_{4}
\end{equation}

\begin{equation}
    L_{0}=\sqrt{r_{2}^2+L_{5}^2-\cos{\alpha_{4}}*2r_{2}L_{5}}
\end{equation}

\begin{equation}
    \alpha_{5}=\arccos{(\frac{L_{0}^2+L_{5}^2-r_{2}^2}{2L_{0}L_{5}})}
\end{equation}

\begin{equation}
    T(x)=\sqrt{3L_{0}^2+3r_{2}^2-\cos{\alpha_{6}}*6r_{2}L_{0}}
\end{equation}

\begin{equation}\label{eq12}
   T_{Aim} =\sqrt{H_{s}^2+(L_{s}-L_{Arc})^2}
\end{equation}

Among Equation~\ref{eq12}, the $H_ {s} $ and $L_ {s} $ are the target height and depth of the stairs, respectively, $T_{Aim}$ stands for the ideal value of $T$. 
Therefore, the $x$, which makes $ T(x) = T_{Aim}$, is the optimal length of the electric push rod.

\section{Experiment and Discussion}
\subsection{Work Scenario Definition}
For SWhegPro, the most significant parameters of terrains are whether the ground is even and the materials of the ground.

But for SWhegPro3, specifically designed for stair climbing, the most considerable parameter is the steepness of the stairs. 
This is quantified by $S$, representing the slope of the stairs, and is defined as follows:

\begin{equation}
    S=arctan(L/H) 
\end{equation}

Where $S$ is measured by degrees, bigger $S$ indicates the stairs are steeper.
$L$ and $H$ represent the length and the height of the stairs. These two parameters are explained in Fig.~\ref{Explain the Parameters of Wheel}(a).

According to the standard for stairs in Chinese civil buildings, the common dimension of stairs can be divided into the following three types:

\begin{itemize}
    \item \textbf{A-type stair:}
    
    $240mm < L < 300mm$, $135mm < H < 165mm$.
    
    \item \textbf{B-type stair:} 
    
    $300mm < L < 450mm$, $135mm < H < 165mm$.
    
    \item \textbf{C-type stair: }
    
    $240mm < L < 300mm$, $80mm < H < 135mm$.
    
\end{itemize}

\begin{table}[htbp]
	\centering
    \caption{Selected Stair Cases}
    \label{Selected stair cases}
	\begin{tabular}{cccc}
		\textbf{Type} & \textbf{Length(mm)} & \textbf{Height(mm)} & \textbf{Slope(\degree)}\\
		\midrule  
             A1 & 260 & 165 & 32.42 \\
             A2 & 300 & 150 & 26.57 \\
             B1 & 340 & 165 & 25.89 \\
             C1 & 300 & 120 & 21.80 \\
             C2 & 300 & 80  & 14.93 \\
		\bottomrule  
	\end{tabular}
\end{table}

We load the simple robots with a series of modes into the Webots and place the physical robots into on-site environments for testing.

For SWhegPro, its work scenarios conclude flat ground, pebbles, grass, and solid uneven ground. Also, we try to climb C-type stairs and analyze its performance.
For SWhegPro3, we can find that A-type stairs are most common in daily life, so we operate the robot on A-type stairs in both simulation environments and fields.
But B-type and C-type stairs are rare in reality, so we test in the Webots simulation environment.

To ensure that our experiments are robust, we randomly chose five non-standard stairs for our test; the parameters of each stair are shown in Table.~\ref{Selected stair cases}.

\subsection{SWhegPro Performance}
\subsubsection{Motion Smoothness on Ground}

Similar to the method used to determine the number of wheels, the motion smoothness of the robot is also evaluated by the cost in the J Function.
Fig.~\ref{Cost of Different Wheel Number}(b) shows the cost of three different modes of SWhegPro when it moves on flat ground.
It can be seen that both RHex mode and SWheg mode have a very small cost, around \ang{0.01}, which means those two modes are suitable for walking on flat ground.

\begin{figure}[t!]
    \centering
    \includegraphics[width=0.9\linewidth, scale=0.8]{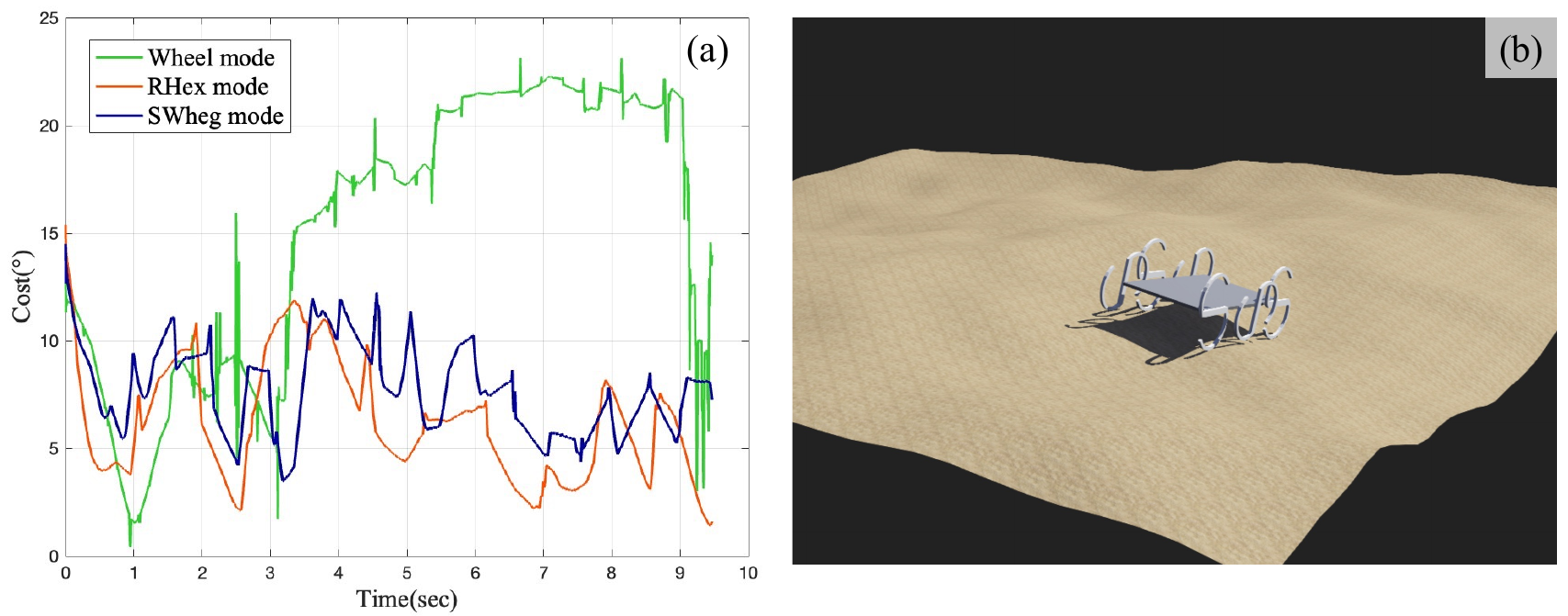}
    \vspace{-1mm}
    \caption{The cost value of SWhegPro under three different modes on the uneven ground (a) and the testing terrain in simulation environment(b).}
    \vspace{-2mm}
    \label{Pro Cost on Uneven Ground}
    \centering
\end{figure}

As can be seen from Fig.~\ref{Pro Cost on Uneven Ground}, when the robot is on terrain with an uneven surface or platform, the RHex mode and the SWheg mode have much better stability than in the wheel mode.

\subsubsection{Stair Test}

\begin{figure}[t!]
    \centering
    \includegraphics[width=0.75\linewidth, scale=0.8]{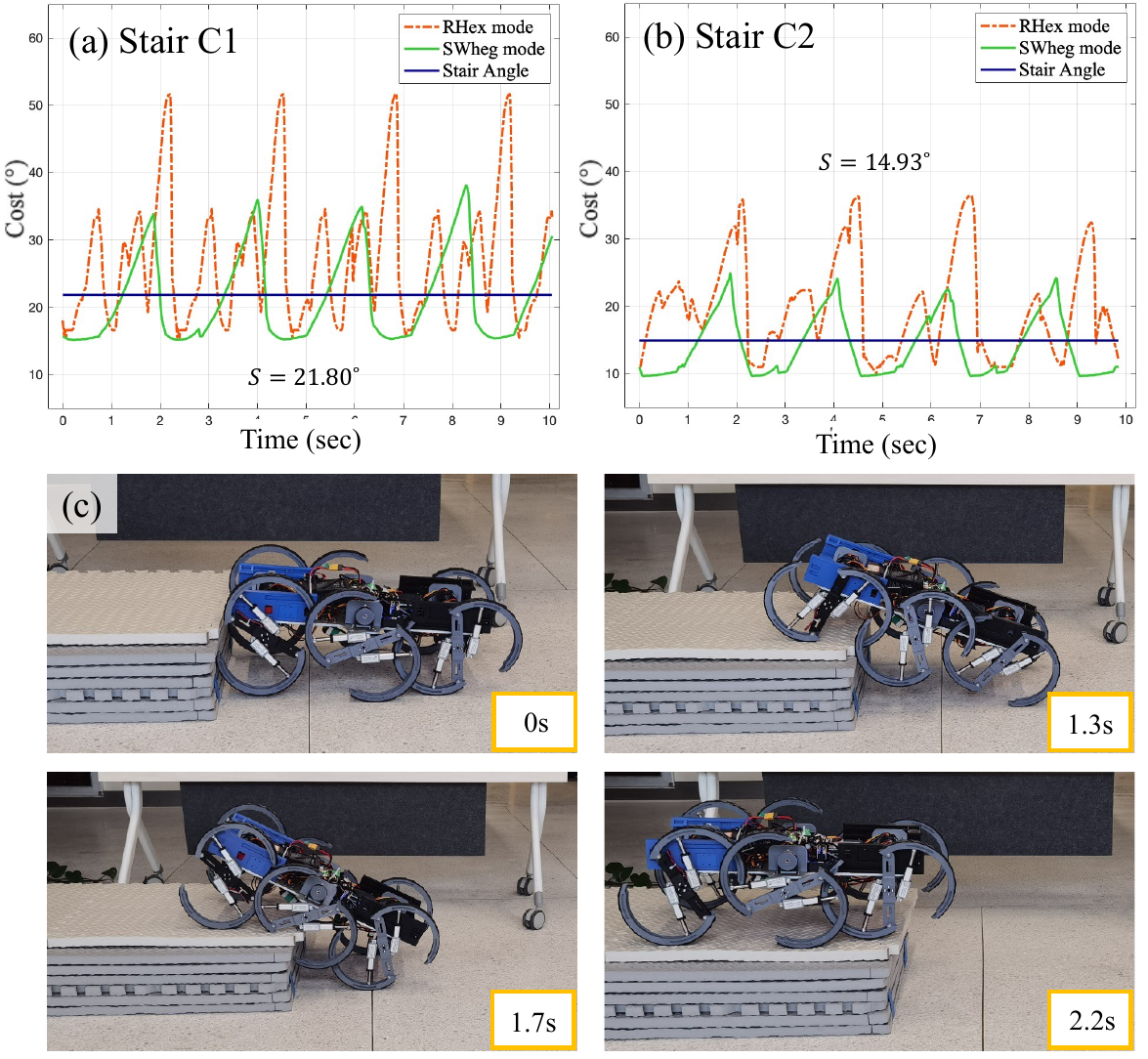}
    \vspace{-1mm}
    \caption{Stair test of SWhegPro. (a) and (b) show the cost of SWhegPro when climbing stairs in the dimensions of $L=300mm,H=120mm$ and $L=300mm, H=80mm$. (c) shows the terrain of maximum step test.}
    \vspace{-2mm}
    \label{Pro Stair Test}
    \centering
\end{figure}

We let SWhegPro climb the stairs in the RHex and SWheg modes to measure its climbing performance and test the maximum height of a step.
Unlike the triangle gait used on flat ground, SWhegPro uses a symmetrical gait to ensure the operation is safe.
Observing the cost shown in Fig.~\ref{Pro Stair Test} (a)(b), it can be found that climbing in the SWheg mode is much more stable than in the RHex mode, but both of them are not stable enough.

Fig.~\ref{Pro Stair Test}(c) shows the max height of the step SWhegPro can step over, which is about 17.5cm.

\subsubsection{Comprehensive Experiment}

\begin{itemize}
    \item 
\textbf{Wheel transformation experiment under load:}
This experiment tests the ultimate load that SWhegPro can hold while operating.
We chose sand as the payload to increase weight, ensuring more precise results continuously. 

According to the self-locked function of the electric push rod, the maximum load is mainly affected by the material and structural strength; as our modular wheels are manufactured by 3D printing tech, the PLA material can't bear too much load. 
So when we ensure the structure is safe, SWhegPro can maintain a normal walking gait with a 12kg load in leg mode and 15kg in wheel mode.

If we promote the mechanical properties of the modular wheels, the maximum load will be limited by the force offered by the actuator, about 60N for each. 
So, the theoretical maximum load of SWhegPro can be expressed by Eq.~\ref{maximum load}. 
Where $L_m$ is the maximum load of SWhegPro, $N$ is the number of wheels, $F$ is the max force of each actuator, and $M$ is the weight of the whole robot.

\begin{equation}
    L_m = \frac{NF}{g} - M \approx 26kg
\label{maximum load}
\end{equation}

\item 
\textbf{Power experiment:}
As shown in Tab.~\ref{tab:ROBOT SPECIFICATIONS}, we also install a power measurement module, hoping to test the energy efficiency of the robot under different ground conditions.

We use the "specific resistance" to evaluate energy efficiency\cite{RN545}.

\begin{equation}
    SR=\frac{P}{Mgv}
\end{equation}

The specific resistance is based on the robot's weight $M$, the average forward speed $v$, and the average power consumption $P$. We measured the power consumption directly from a power measurement module.

Table~\ref{Pro Power} shows the average forward speed, power, and the specific resistance of 5 runs over flat ground, grass, and pebbles for all tests. Some conclusions can be derived from Table~\ref{Pro Power}:

\begin{enumerate}
\item 
SWhegPro moves well over three terrains, with speed variations range: $0.213m/s \sim 0.332m/s$ for wheeled mode and $0.196m/s \sim 0.282m/s$ for legged mode(tripod gait).
In both modes, the forward speed over pebbles is the lowest due to the pebble terrain's discontinuity and slippery characteristics. The speed variation range of the legged mode is smaller than that of the wheeled mode because the discontinuity contact of the legged mode decreases the influence of terrains.

\item
In both modes over three terrains, SWhegPro always consumes more energy at a faster speed. And it is more energy efficient while traversing flat ground compared to grass and pebbles. Moreover, SWhegPro features a lower SR value in wheeled mode compared to legged locomotion at a similar speed. The three terrains we selected in this experiment can all be classified into flat terrain categories, compared to steps and stairs being rough terrains. This shows that wheels are more energy efficient on flat terrains, hence the ideal strategy for traversing flat terrains.
\end{enumerate}

\end{itemize}

\begin{table}[h]
\centering  
\renewcommand\arraystretch{1.5}
\footnotesize
\caption{Velocity and Energy Efficiency Comparison}
\label{Pro Power}
\begin{tabular}{clcccc}
\textbf{Mode} & \textbf{Features} & \textbf{Flat ground} & \textbf{Pebbles} & \textbf{Grass} \\ \hline
              & velocity (m/s)    & 0.332                & 0.213            & 0.287          \\
Wheeled mode  & power (W)         & 20.475               & 13.423           & 17.857         \\
              & SR                & 0.623                & 0.638            & 0.629          \\ \hline
              & velocity (m/s)    & 0.282                & 0.196            & 0.249          \\
Legged mode   & power (W)         & 18.640               & 15.657           & 18.354         \\
              & SR                & 0.667                & 0.808            & 0.746          \\ \hline
\end{tabular}
\end{table}

\subsection{SWhegPro3 Performance}

\subsubsection{Moving Smoothness}

\begin{figure*}[t!]
    \centering
    \includegraphics[width=0.98\linewidth, scale=0.8]{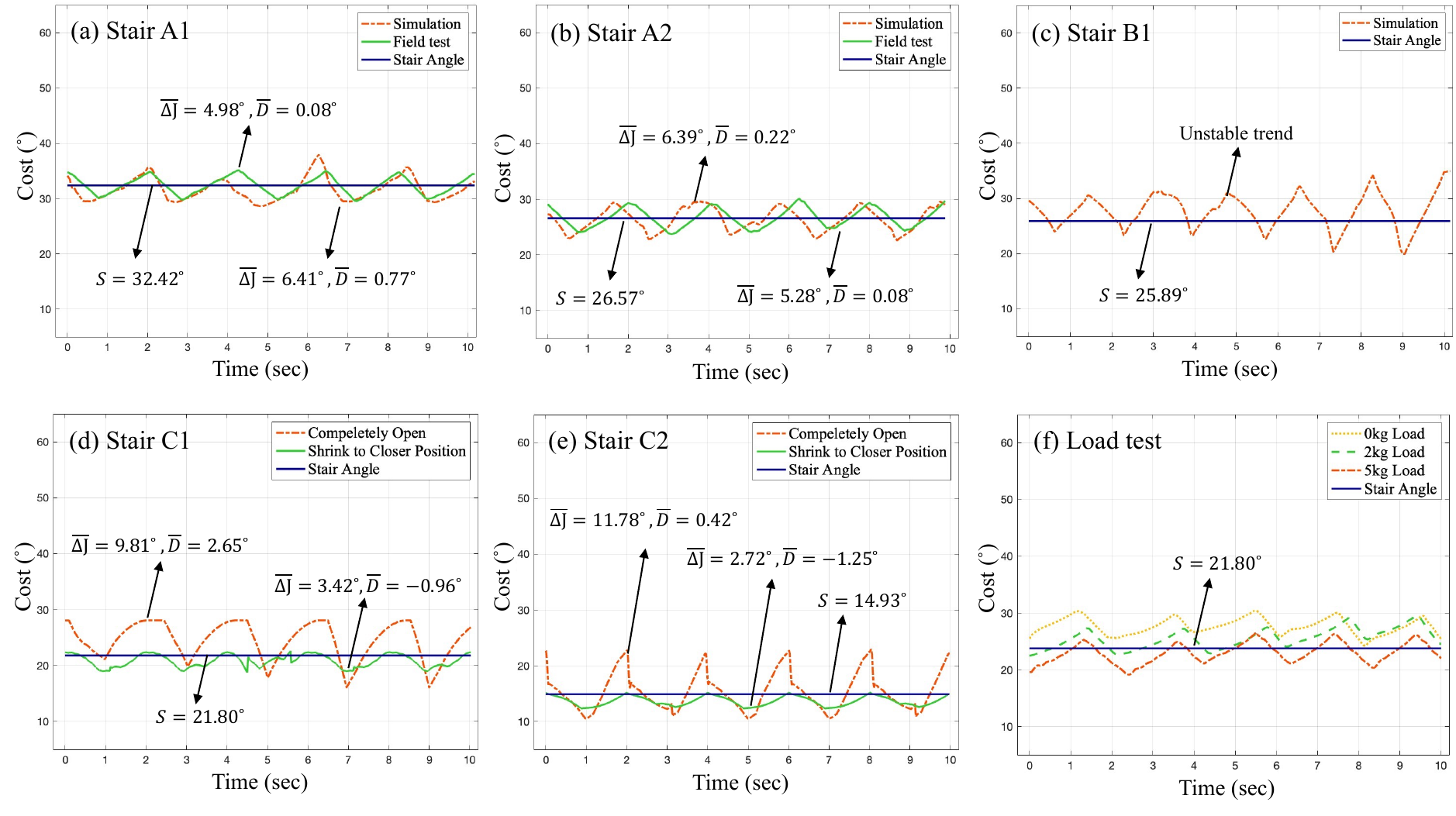}
    \vspace{-1mm}
    \caption{Results of SWhegPro3’s stair tests. (a) and (b) show the cost while stably climbing stairs A1 \& A2. (c) shows the unstable climbing condition on stair B1. (d) and (e) is the verification of the optimal method according to climb on stairs C1 \& C2. (f) is the result of the load test of SWhegPro3.}
    \vspace{-2mm}
    \label{Pro3 Stair Test}
    \centering
\end{figure*}  

To capture the robot's motion performance, we record IMU data from five periods after reaching a stable climbing state. 
During the analysis, we define the following two parameters to measure its performance. 
$\Bar{D}$ indicates the average deviation between the cost $J$ and the stair slope angle $S$, which can be used to measure the risk of tipping over, expressed by Eq.~\ref{Deviation}.
$\overline{\Delta J}$ indicates the average gap, which can directly measure the stability of the robot, expressed by Eq.~\ref{Delta J}.

\begin{equation}
\label{Deviation}
    \overline{D}=\frac{\sum \limits_{i=0}^{T-1} (J_i-S)}{T}
\end{equation}

Where $T$ is the number of timestamp, $S$ is the slope of stair refers from Table.~\ref{Selected stair cases}

\begin{equation}
\label{Delta J}
    \overline{\Delta J}=\frac{\sum \limits_{i=1}^N (J^{max}_i-J^{min}_i)}{N}
\end{equation}

Where $J^{max}_i$ and $J^{min}_i$ indicate each cycle's local maximum and minimum point, $N$ is the number of periods.

\begin{figure}[t!]
    \centering
    \includegraphics[width=0.9\linewidth, scale=0.8]{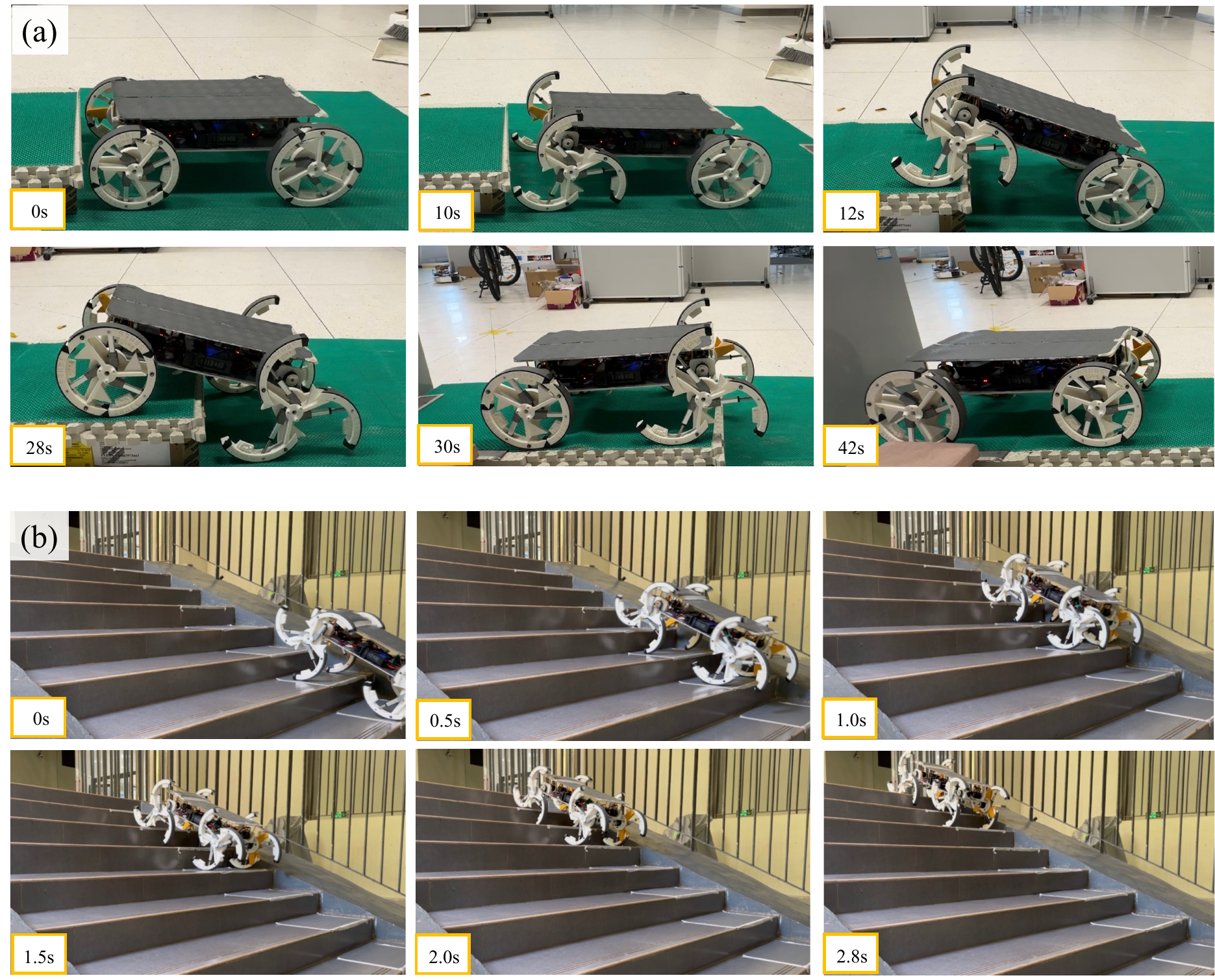}
    \vspace{-1mm}
    \caption{SWhegPro3's comprehensive test. Test of highest single step in adjustable terrain (a) and the maximum climbing speed at stair C1 (b).}
    \vspace{-2mm}
    \label{Pro3 Comprehensive Test}
    \centering
\end{figure} 

While SWhegPro3 climbing stair A1 in simulation environment, the average deviation $\overline{D_{A1}^{Sim}}=\ang{0.77}(\approx2.38\%)$, and the average maximum difference $\overline{{\Delta J}_{A1}^{Sim}} = \ang{6.41}(\approx19.77\%)$.
And in the field test, the average deviation $\overline{D_{A1}^{Field}}=\ang{0.08}(\approx0.25\%)$, the average maximum difference $\overline{{\Delta J}_{A1}^{Field}} = \ang{4.98}(\approx15.36\%)$.

When the robot climbs stair A2 in simulation environment, the average deviation $\overline{D_{A2}^{Sim}}=\ang{0.22}(\approx0.83\%)$, and the average maximum difference $\overline{{\Delta J}_{A2}^{Sim}} = \ang{6.39}(\approx24.25\%)$.
In the field test, the average deviation $\overline{D_{A2}^{Field}}=\ang{0.08}(\approx0.3\%)$, the average maximum difference $\overline{{\Delta J}_{A2}^{Field}} = \ang{5.28}(\approx19.87\%)$.

According to a rapid glance at the results shown in Fig .~\ref{Pro3 Stair Test}(a)(b), we can determine that the performance of SWhegPro3 climbing common stairs is outstanding.
Also, we observe an interesting phenomenon: both the $\overline{D}$ and the $\overline{\Delta J}$ in the field test are lower than in simulation, this might due to the excessive output in simulation.

When SWhegPro3 climbs on uncommon stair B1, the length of spokes is insufficient: the contact point between the spokes and the stairs will be closer and closer to the spoke's end, after climbing certain steps, the spokes will miss a step and collide to the stair surface, leading to a pronounced drop in climbing stability. 
According to the cost curve shown in Fig.~ \ref{Pro3 Stair Test}(c), we deduce that the operation of SWhegPro3 on stair B1 is not stable.

\subsubsection{Verification of Transformation Controlling Method}

According to the calculation mentioned in wheel module control, we find that while the robot climbs the stairs whose height is not lower than 135mm, adjusting the modular wheel cannot acquire prominent promotion of stability, so the control strategy is only used in C-type stairs.

To validate the effect of our mathematical model, we simulate the robot in two C-type stairs mentioned in Table.~\ref{Selected stair cases}, and load the optimal length $T_{Aim}$ mentioned in Eq.~\ref{eq12}.

So according to the data shown in Fig.~\ref{Pro3 Stair Test}(d)(e), when SWhegPro3 with the origin completely opened wheel climbs the stair C1, the average deviation $\overline{D_{C1}^{Origin}}=\ang{2.65}(\approx12.15\%)$, and the average maximum difference $\overline{{\Delta J}_{C1}^{Origin}} = \ang{9.81}(\approx45\%)$; while climbing with the optimal shrink wheel module, the average deviation $\overline{D_{C1}^{Optimal}}=\ang{-0.96}(\approx-4.39\%)$, the average maximum difference $\overline{{\Delta J}_{C1}^{Optimal}} = \ang{3.42}(\approx15.68\%)$.

When SWhegPro3 with the origin completely opened wheel climbs the stair C2, the average deviation $\overline{D_{C2}^{Origin}}=\ang{0.42}(\approx2.78\%)$, and the average maximum difference $\overline{{\Delta J}_{C2}^{Origin}} = \ang{11.78}(\approx78.89\%)$; while climbing with the optimal shrink wheel module, the average deviation $\overline{D_{C2}^{Optimal}}=\ang{-1.25}(\approx-8.37\%)$, the average maximum difference $\overline{{\Delta J}_{C2}^{Optimal}} = \ang{2.72}(\approx18.21\%)$.

According to the definition of $\overline{D}$ and $\overline{\Delta J}$, we can find that SWhegPro3 with optimal shrink wheel has a much lower risk of tipping over and significant improvement($>300\%$) of smoothness and stability, which means the wheel module optimal method is effective.

\subsubsection{Comprehensive Performance Experiments}
\begin{itemize}
    \item 
Maximum transportable load：
We observe substantial horizontal drift when the robot climbs on stair C1 without any load, 
$\overline{D}=\ang{3.83} (16.09\%)$. 
This discrepancy implies that the unexpected torque output between the front and rear wheels significantly increases the risk of tipping over.
Thus, we put a 2kg load at the front of the robot for stabilization, and Fig.~\ref{Pro3 Stair Test}(f) illustrates the cost variation while applying different loads.

When the external load is 2kg, the average deviation $\overline{D}=\ang{1.82}(7.65\%)$; when the external load is 5kg, the average deviation $\overline{D}=\ang{0.82}(3.45\%)$. 
It can be inferred that the risk of tipping is lower when the robot has a certain load on the forward.
Simultaneously, we realize a stable climbing with an external load of 7kg, as the maximum load of SWhegPro3. 
Heavier external loads may exhibit slippery of the belt, which is unsafe.
    \item 
Highest single step： 
To determine the maximum height that the robot can climb, we construct a step whose height can change by using foam pads, with each pad being 2cm thick.
As shown in Fig.~\ref{Pro3 Comprehensive Test}(a), the robot can stably climbs over the 20cm-height stair.
The robot cannot climb higher in a single step is because the bigger pitch angle, which makes the front wheel lack friction, and the torque output of the rear wheel insufficient.
    \item 
Maximum climbing speed： 
Considering the potential tipping hazard while performing the maximum speed test on a steeper staircase, the maximum tests are carried out on C1-type stairs.

Shown in Fig.~\ref{Pro3 Comprehensive Test}(b), the robot can climb five stairs in a mere 5 seconds, translating to approximately 1.8 steps per second.
When attempting even higher speeds, we encounter a pivotal constraint: higher target speed leads to excessive torque output, causing the robot to tip over or slip. 
Consequently, strategies such as redistributing the body's weight and enhancing the spokes' friction-generating capability are considered avenues to increase the robot's maximum climbing speed. 
\end{itemize}

\section{Conclusion}
In this article, we proposed the design principle for a transformation mechanism in a wheel-leg transformable robot. Furthermore, we designed and tested two robots for different environments, SWhegPro(outdoor environment) and SWhegPro3(Indoor environment). The wheel-leg transformation mechanism is designed with the same principle but has different shapes suitable for each environment. The design parameter selection is also driven by other use cases and validated with extensive smoothness and traversability experiments. 

Future improvement can be found in adding perception capabilities that lead to better high-level decision-making. Second, Improve the modality of the wheel-leg transformable module to allow quick change when functioning in different environments.

\bibliographystyle{ieeetr}
\bibliography{references}

\begin{thebibliography}{10}

\bibitem{kim2014wheel}
Y.-S. Kim, G.-P. Jung, H.~Kim, K.-J. Cho, and C.-N. Chu, ``Wheel transformer: A wheel-leg hybrid robot with passive transformable wheels,'' {\em IEEE Transactions on Robotics}, vol.~30, no.~6, pp.~1487--1498, 2014.

\bibitem{cao2022omniwheg}
R.~Cao, J.~Gu, C.~Yu, and A.~Rosendo, ``Omniwheg: An omnidirectional wheel-leg transformable robot,'' {\em arXiv preprint arXiv:2203.02118}, 2022.

\bibitem{chen2014Quattroped}
S.-C. Chen, K.-J. Huang, W.-H. Chen, S.-Y. Shen, C.-H. Li, and P.-C. Lin, ``Quattroped: A leg--wheel transformable robot,'' {\em IEEE/ASME Transactions on Mechatronics}, vol.~19, no.~2, pp.~730--742, 2014.

\bibitem{sun2017transformable}
T.~Sun, X.~Xiang, W.~Su, H.~Wu, and Y.~Song, ``A transformable wheel-legged mobile robot: Design, analysis and experiment,'' {\em Robotics and Autonomous Systems}, vol.~98, pp.~30--41, 2017.

\bibitem{Bai2018An}
L.~Bai, J.~Guan, X.~Chen, J.~Hou, and W.~Duan, ``An optional passive/active transformable wheel-legged mobility concept for search and rescue robots,'' {\em Robotics and Autonomous Systems}, vol.~107, pp.~145--155, 2018.

\bibitem{Marques2006RAPOSA}
C.~F. Marques, J.~Cristovao, P.~U. Lima, J.~Frazao, and R.~Ventura, ``Raposa: Semi-autonomous robot for rescue operations,'' in {\em 2006 IEEE/RSJ International Conference on Intelligent Robots and Systems, IROS 2006, October 9-15, 2006, Beijing, China}, 2006.

\bibitem{Saranli2001RHex}
U.~Saranli, ``Rhex: A simple and highly mobile hexapod robot,'' {\em The International Journal of Robotics Research}, vol.~20, no.~7, pp.~616--631, 2001.

\bibitem{agrawal2016ions}
S.~P. Agrawal, H.~Dagale, N.~Mohan, and L.~Umanand, ``Ions: a quadruped robot for multi-terrain applications,'' {\em Int. J. Mater. Mech. Manuf.}, vol.~4, pp.~84--88, 2016.

\bibitem{Nakajima2004Motion}
S.~Nakajima, E.~Nakano, and T.~Takahashi, ``Motion control technique for practical use of a leg-wheel robot on unknown outdoor rough terrains,'' in {\em International Conference on Intelligent Robots and Systems}, 2004.

\bibitem{Kosugi1984Motion}
M.~Kosugi, T.~Ohya, and T.~Migita, ``Motion analysis with experimental verification of the hybrid robot peopler-ii for reversible switch between walk and roll on demand.,'' {\em Chemischer Informationsdienst}, vol.~15, no.~11, pp.~no--no, 1984.

\bibitem{Lacagnina2003Kinematics}
M.~Lacagnina, G.~Muscato, and R.~Sinatra, ``Kinematics, dynamics and control of a hybrid robot wheeleg,'' {\em Robotics and Autonomous Systems}, vol.~45, no.~3-4, pp.~161--180, 2003.

\bibitem{WLinARM}
L.~Zhengtao, D.~Cunxi, L.~Xiaohan, Z.~Jianxiang, and J.~Zhenzhong, ``A hybrid wheel-leg transformable robot with minimal actuator realization,'' in {\em 2013 IEEE International Conference on Advanced Robotics and Mechatronics}, pp.~5625--5630, 2013.

\bibitem{ryu2020shape}
S.~Ryu, Y.~Lee, and T.~Seo, ``Shape-morphing wheel design and analysis for step climbing in high speed locomotion,'' {\em IEEE Robotics and Automation Letters}, vol.~5, no.~2, pp.~1977--1982, 2020.

\bibitem{Lee2017Origami}
D.~Y. Lee, S.~R. Kim, J.~S. Kim, J.~J. Park, and K.~J. Cho, ``Origami wheel transformer: A variable-diameter wheel drive robot using an origami structure,'' {\em Soft Robotics}, vol.~4, no.~2, pp.~163--180, 2017.

\bibitem{Moriya2020Robotic}
N.~Moriya, H.~Shigemune, and H.~Sawada, ``A robotic wheel locally transforming the diameter for the locomotion on rough terrain,'' in {\em 2020 IEEE International Conference on Mechatronics and Automation (ICMA)}, pp.~1257--1262, 2020.

\bibitem{R2020FUHAR}
R.~Mertyüz, A.~K. Tanyldz, B.~Taar, A.~B. Tatar, and O.~Yakut, ``Fuhar: A transformable wheel-legged hybrid mobile robot,'' {\em Robotics and Autonomous Systems}, vol.~133, 2020.

\bibitem{chen2013quattroped}
S.-C. Chen, K.-J. Huang, W.-H. Chen, S.-Y. Shen, C.-H. Li, and P.-C. Lin, ``Quattroped: a leg--wheel transformable robot,'' {\em IEEE/ASME Transactions On Mechatronics}, vol.~19, no.~2, pp.~730--742, 2013.

\bibitem{kim2020step}
Y.~Kim, Y.~Lee, S.~Lee, J.~Kim, H.~S. Kim, and T.~Seo, ``Step: A new mobile platform with 2-dof transformable wheels for service robots,'' {\em IEEE/ASME Transactions on Mechatronics}, vol.~25, no.~4, pp.~1859--1868, 2020.

\bibitem{won_design_2022}
J.~Won, S.~Ryu, S.~Kim, K.~Y. Yoo, H.~S. Kim, and T.~Seo, ``Design optimization of a linkage-based 2-{DOF} wheel mechanism for stable step climbing,'' {\em Scientific Reports}, vol.~12, p.~16912, Oct. 2022.

\bibitem{dai2022swheg}
C.~Dai, X.~Liu, J.~Zhou, Z.~Liu, and Z.~Jia, ``Swheg: A wheel-leg transformable robot with minimalist actuator realization,'' {\em arXiv preprint arXiv:2210.15126}, 2022.

\bibitem{dai2022swhegpro}
C.~Dai, X.~Liu, J.~Zhou, Z.~Liu, Z.~Zhu, and Z.~Jia, ``Swhegpro: A novel robust wheel-leg transformable robot,'' in {\em 2022 IEEE International Conference on Robotics and Biomimetics (ROBIO)}, pp.~421--426, IEEE, 2022.

\bibitem{wang2023swhegpro3}
H.~Wang, S.~Wang, C.~Dai, and Z.~Jia, ``Swhegpro3: A three-impeller wheel-leg transformable robot with variable robust adaptability to stair dimensions,'' in {\em 2023 IEEE International Conference on Robotics and Biomimetics (ROBIO)}, pp.~1--6, IEEE, 2023.

\bibitem{chen2017turboquad}
W.-H. Chen, H.-S. Lin, Y.-M. Lin, and P.-C. Lin, ``Turboquad: A novel leg--wheel transformable robot with smooth and fast behavioral transitions,'' {\em IEEE Transactions on Robotics}, vol.~33, no.~5, pp.~1025--1040, 2017.

\bibitem{kim2019curved}
Y.~Kim, J.~Kim, H.~S. Kim, and T.~Seo, ``Curved-spoke tri-wheel mechanism for fast stair-climbing,'' {\em IEEE Access}, vol.~7, pp.~173766--173773, 2019.

\bibitem{8793686}
C.~Zheng and K.~Lee, ``Wheeler: Wheel-leg reconfigurable mechanism with passive gears for mobile robot applications,'' in {\em 2019 International Conference on Robotics and Automation (ICRA)}, pp.~9292--9298, 2019.

\bibitem{RN545}
G.~V.~T. Gabrielli, {\em What price speed? : specific power required for propulsion of vehicles}.
\newblock 1950.

\end{thebibliography}

\end{CJK}
\end{document}